%% file: main.tex
\pdfoutput=1

\documentclass[11pt]{article}

\usepackage[final]{acl}

\input{sections/preamble}

\title{Token-Budget-Aware LLM Reasoning}

\author{\bf Tingxu Han\footnotemark[1]$^1$, \bf Zhenting Wang\footnotemark[1]\footnotemark[2]$^2$, \bf Chunrong Fang\footnotemark[3]$^1$,\\
\bf Shiyu Zhao$^2$, \bf Shiqing Ma$^3$, \bf Zhenyu Chen$^1$\\
$^{1}$Nanjing University
$^{2}$Rutgers University
$^{3}$UMass Amherst
}

\begin{document}
\maketitle
{
\renewcommand{\thefootnote}{\fnsymbol{footnote}}
\footnotetext[1]{Equal Contribution.}
\footnotetext[2]{Start the project and propose the idea.}
\footnotetext[3]{Corresponding Author.}
}
\input{sections/abstract}

\input{sections/introduction}

\input{sections/related_work}

\input{sections/observation}

\input{sections/methodology}

\input{sections/evaluation}

\input{sections/conclusion}

\input{sections/limitation}

\bibliography{refer}
\input{sections/appendix}

\end{document}

%% file: sections/preamble.tex
\usepackage{times}
\usepackage{latexsym}

\usepackage[T1]{fontenc}

\usepackage[utf8]{inputenc}

\usepackage{microtype}
\usepackage{amsmath}
\usepackage{booktabs}
\usepackage{subfig}
\usepackage{array}    %
\usepackage{ragged2e}
\usepackage{multicol}
\usepackage{multirow}
\usepackage{inconsolata}
\usepackage{makecell}
\usepackage{graphicx}

\newcommand{\RomanOne}{\uppercase\expandafter{\romannumeral 1} }
\newcommand{\RomanTwo}{\uppercase\expandafter{\romannumeral 2} }
\newcommand{\RomanThree}{\uppercase\expandafter{\romannumeral 3} }
\newcommand{\RomanFour}{\uppercase\expandafter{\romannumeral 4} }

\newcommand{\ours}{\textsc{TALE}}

\newcommand{\revise}[1]{{\color{black}{#1}}}

\definecolor{cvprblue}{rgb}{0.21,0.49,0.74}

\newcommand{\sys}{\mbox{\textsc{TALE }}}

\usepackage{algorithm}
\usepackage{algorithmicx}
\usepackage[noend]{algpseudocode}

\input{utils/math_commands}

%% file: utils/math_commands.tex
\usepackage{amsmath,amsfonts,bm}

\def\eqref#1{equation~\ref{#1}}

\def\1{\bm{1}}

\def\vx{{\bm{x}}}

\DeclareMathAlphabet{\mathsfit}{\encodingdefault}{\sfdefault}{m}{sl}
\SetMathAlphabet{\mathsfit}{bold}{\encodingdefault}{\sfdefault}{bx}{n}

%% file: sections/abstract.tex
\begin{abstract}

Reasoning is critical for large language models (LLMs) to excel in a wide range of tasks. While methods like Chain-of-Thought (CoT) reasoning and enhance LLM performance by decomposing problems into intermediate steps, they also incur significant overhead in token usage, leading to increased costs. 
We find that the reasoning process of current LLMs is unnecessarily lengthy and it can be compressed by including a reasonable token budget in the prompt, but the choice of token budget plays a crucial role in
the actual compression effectiveness. 
We then propose a token-budget-aware LLM reasoning framework that dynamically adjusts the number of reasoning tokens based on the reasoning complexity of each problem. Experiments show that our method effectively reduces token costs in CoT reasoning with only a slight performance reduction, offering a practical solution to balance efficiency and accuracy in LLM reasoning. Code: 
\url{https://github.com/ControllableModel/TALE}\footnote{Also available at \url{https://www.gitlink.org.cn/txhan/TALE}}.

\end{abstract}

%% file: sections/introduction.tex
\begin{verse}
    \emph{``It is not enough to have a good mind; the main thing is to use it well.''}

    \hfill $-$ René Descartes
    
\end{verse}

\vspace{-0.1cm}
\section{Introduction}
\label{sec:intro}
\vspace{-0.1cm}

\begin{figure}[!t]
\centering
     \subfloat[Direct answering (15 output tokens).]
    {
       \includegraphics[width=0.83\columnwidth]{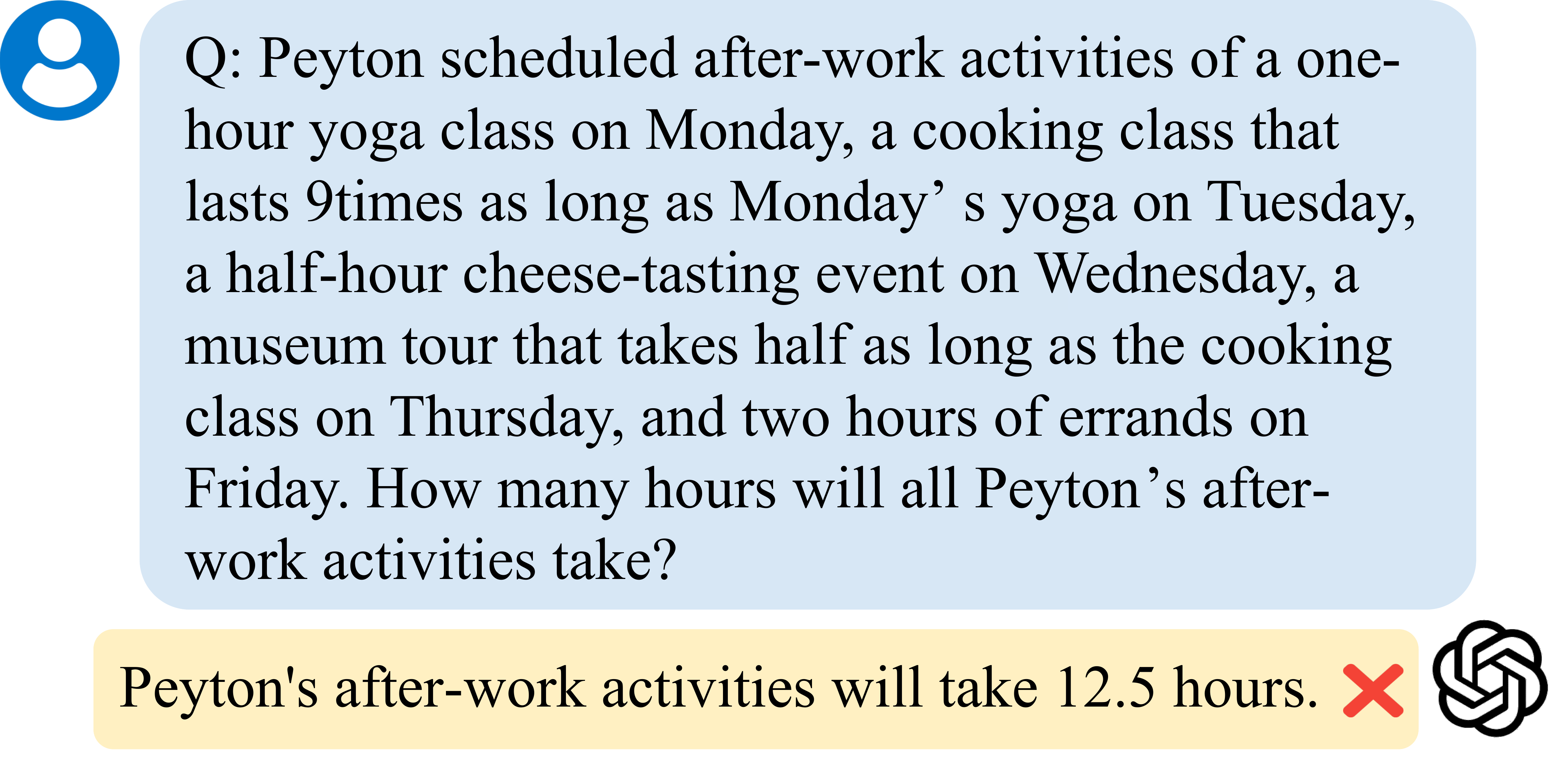}
       \label{fig:case_no_CoT}
    }
    \vfill
    \vspace{-8pt}
    \subfloat[Vanilla CoT (258 output tokens).]
    {
        \includegraphics[width=0.83\columnwidth]{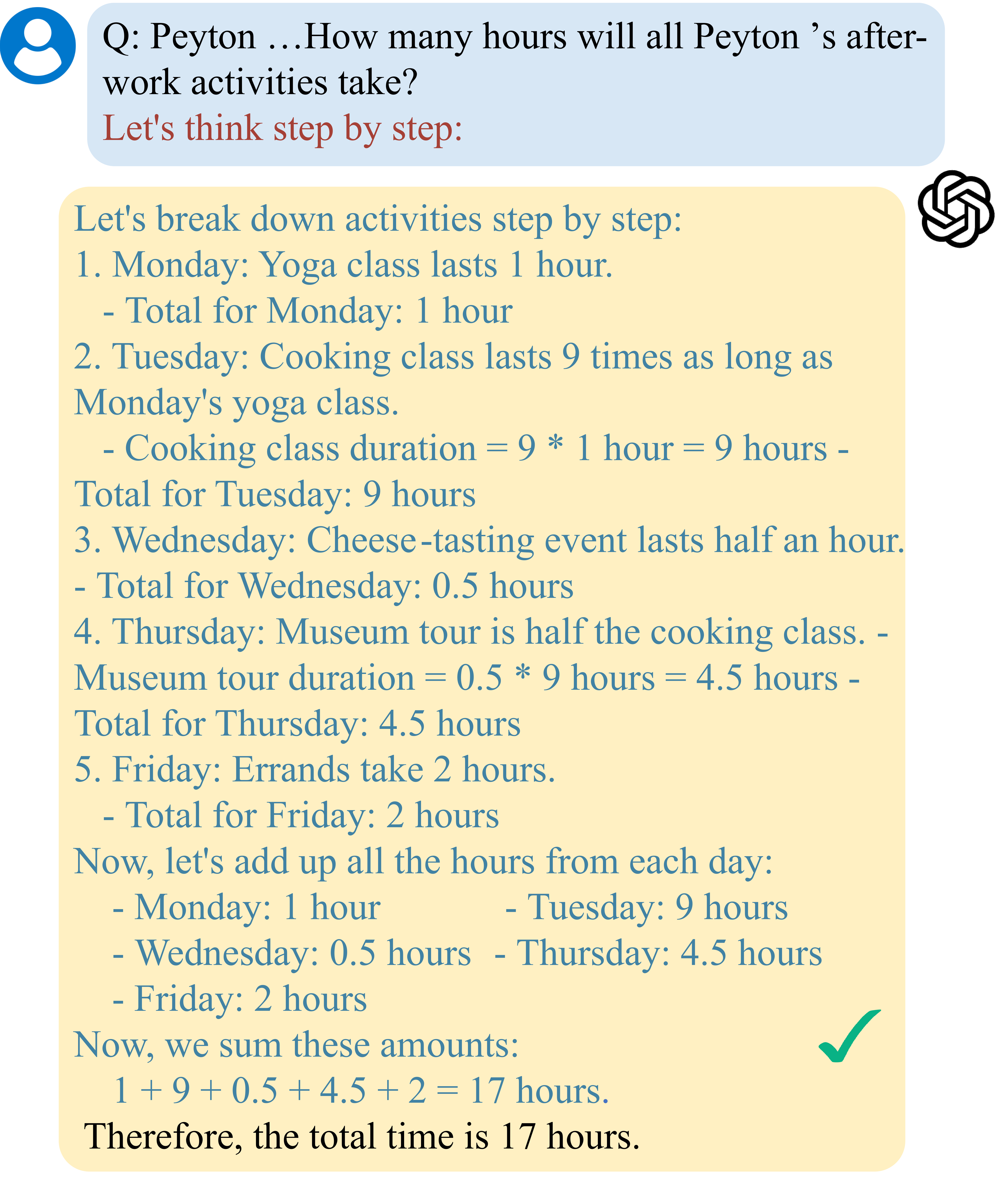}
        \label{fig:case_vanilla_CoT}
    }
    \vfill
    \vspace{-8pt}
    \subfloat[CoT with an unreasonable budget (157 output tokens).]
    {
        \includegraphics[width=0.83\columnwidth]{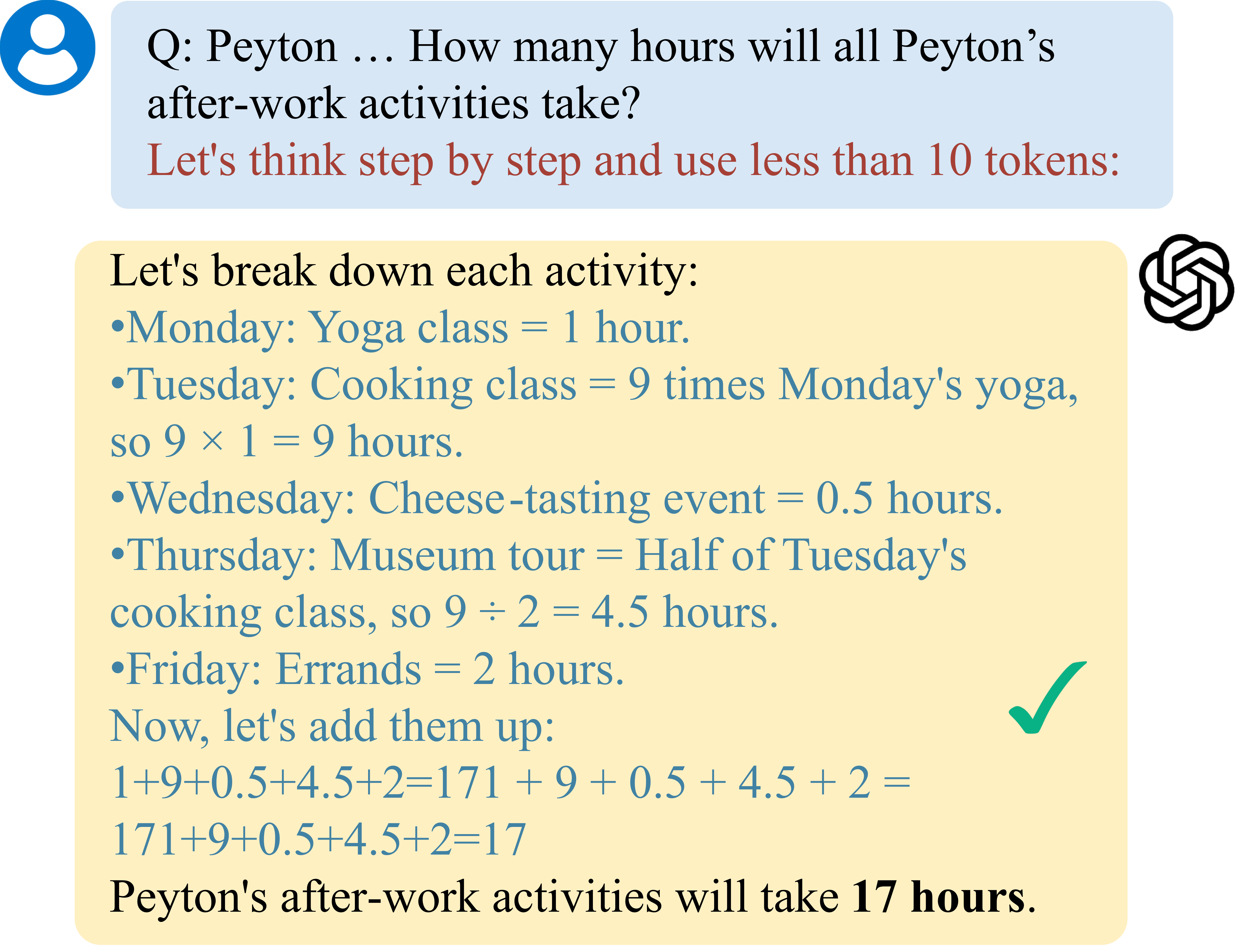}
        \label{fig:one-budget}
    }
    \vfill
    \vspace{-8pt}
    \subfloat[CoT with an reasonable budget (86 output tokens).]
    {
        \includegraphics[width=0.83\columnwidth]{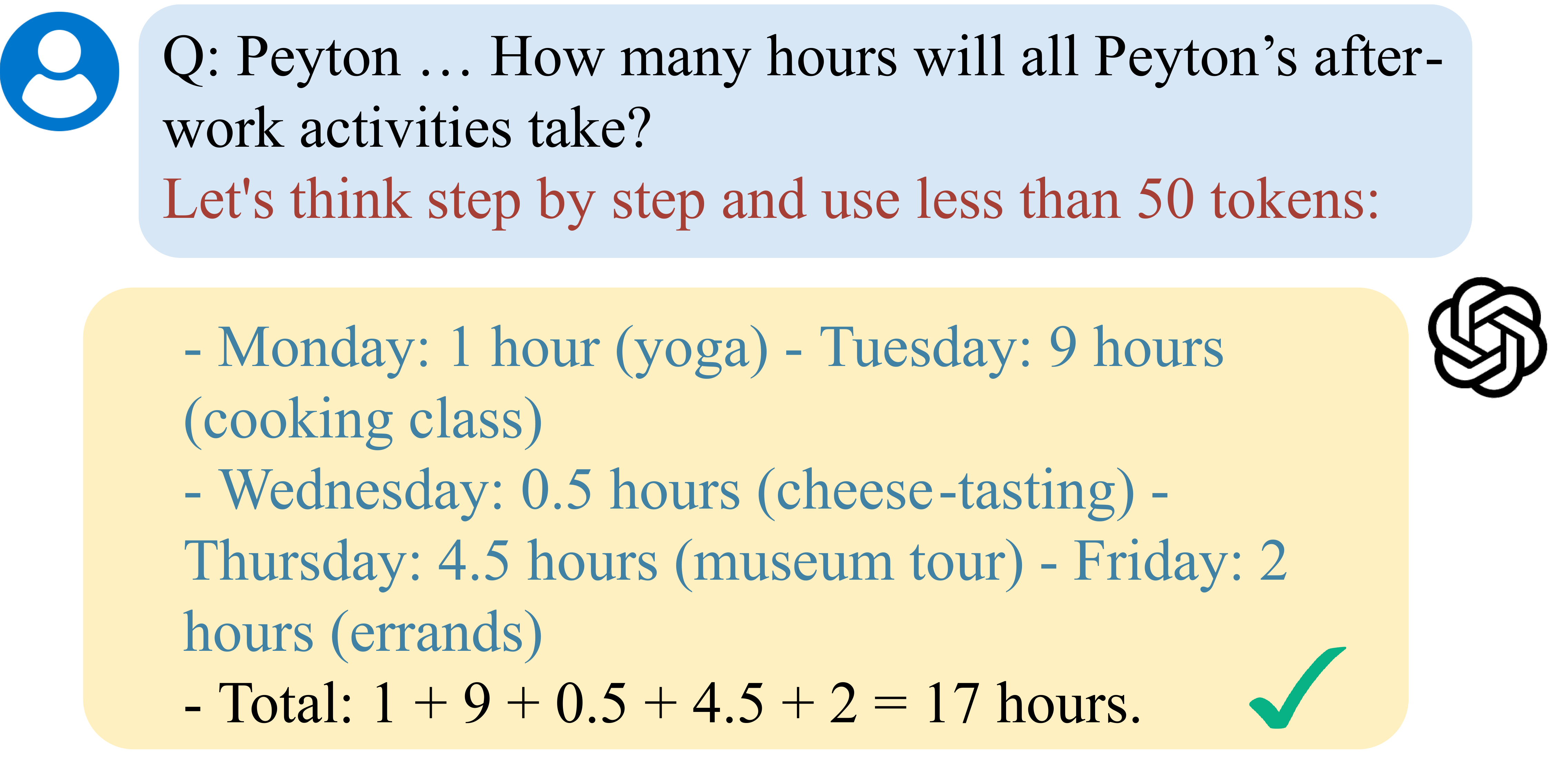}
        \label{fig:case_budget_CoT}
    }
    \vspace{-6pt}
    \caption{Examples of different problem solving paradigms. The reasoning processes are highlighted. We conduct this experiment on GPT-4o-mini.}
    \label{fig:case_study_cot}
    \vspace{-16pt}
\end{figure}

Reasoning plays a crucial role in enabling large language models (LLM) to perform effectively across a wide range of tasks~\cite{zhou2022least,hao2023reasoning,2024-llmreasoner,jin2024exploring,wang2024mllm,wang2025dump}.
A variety of methods have been proposed to enhance the reasoning capabilities of large language models~\cite{suzgun2022challenging,wang2023plan,feng2023alphazero,xie2024monte}. Among these, Chain-of-Thought (CoT)~\cite{2022-CoT} is the most representative and widely adopted approach. 
It enhances the reliability of the model's answers by guiding large language models with the prompt ``Let's think step by step'', encouraging them to decompose the problem into intermediate steps and solve each before arriving at the final answer. 
\autoref{fig:case_no_CoT} and \autoref{fig:case_vanilla_CoT} illustrate an intuitive example.
Observe that without CoT,  the LLM produces incorrect answers to the question. 
With a CoT-enhanced prompt, the LLM systematically breaks the question into multiple steps and reasons through each step sequentially. 
By addressing each step incrementally, the LLM eventually arrives at the correct answer. 
Recent reasoning models, such as OpenAI O1~\cite{o1} and DeepSeek R1~\cite{guo2025deepseek}, integrate CoT into their design. Notably, these models can perform CoT reasoning even without explicit prompting.

Although reasoning enhancement approaches such as CoT impressively improve LLM performance, they produce substantial additional overhead, specifically in the form of the increased number of tokens produced~\cite{2022-CoT,feng2023alphazero,yao2024tree,jin2024impact}.
As shown in \autoref{fig:case_vanilla_CoT}, the answer to prompt with CoT has notably higher token costs due to the detailed intermediate reasoning steps included in the output.
Such high token costs can lead to significant expenses, including increased computational resource usage and longer running times during the LLM inference, ultimately resulting in significant additional monetary and energy costs.

This raises an important question: \emph{``Is the reasoning process of current LLMs unnecessarily lengthy, and how can it be compressed?''} 
\citet{2024-CCoT} demonstrate that LLM has the potential to follow a length constraint in the prompt. 
Building on this, we find that \emph{including a \textbf{token budget} (see \autoref{tab:prompt_construction}) in the prompts is a promising approach to compressing the CoT reasoning tokens. However, the choice of token budget plays a crucial role in the actual compression effectiveness.}
For example, \autoref{fig:case_budget_CoT} illustrates that including a reasonable token budget (e.g., 50 tokens in this case) in the instructions reduces the token cost in the chain-of-thought (CoT) process from 258 output tokens to 86 output tokens, while still enabling the LLM to arrive at the correct answer. However, when the token budget is set to a different smaller value (e.g., 10 tokens), the output token reduction is less effective, resulting in 157 output tokens—nearly twice as many as with a 50-token budget.
In other words, when the token budget is relatively small, LLMs often fail to follow the given token budget. In such cases, the actual token usage significantly exceeds the given budget—even much larger than the token costs with larger token budgets. We refer to this phenomenon as the ``Token Elasticity'' in the CoT process with token budgeting.
To address this, the optimal token budget for a specific LLM and a particular question can be searched by gradually reducing the budget specified in the prompt, identifying the smallest token budget that achieves both the correct answer and the lowest actual token cost.

Based on the above observations and analysis, we propose a token-budget-aware LLM reasoning framework that dynamically adjusts the number of reasoning tokens based on the reasoning complexity of each problem.
We call our method \ours{} (\underline{T}oken-Budget-\underline{A}ware \underline{L}LM r\underline{E}asoning), which includes two implementations: token budget estimation and prompting (\ours{}-EP) and token budget awareness internalization via post-training (\ours{}-PT).
\ours{}-EP estimates a reasonable token budget for each problem using zero-shot prompting and incorporates it into the reasoning process, while \ours{}-PT internalizes token-budget awareness through post-training, enabling the LLM to generate more token-efficient responses without explicit token constraints in the prompt. We discuss both implementations in \autoref{sec:meth}. Experiment results show that \ours{} significantly reduces token costs in LLM chain-of-thought (CoT) reasoning while largely maintaining answer correctness. On average, \ours{}-EP achieves a 67\% reduction in token usage while maintaining accuracy with less than a 3\% decrease.
\ours{}-PT cuts token usage by around 50\% compared to Vanilla CoT and achieves competitive performance.

%% file: sections/related_work.tex
\vspace{-0.1cm}
\section{Related Work}
\label{sec:related_work}
\vspace{-0.1cm}
\noindent\textbf{LLM Reasoning.}
Reasoning in LLMs has seen substantial advancements through techniques that generate intermediate steps, enabling more accurate and effective performance across diverse domains~\cite{wu2022ai,yang2022seqzero,zhou2022least,sun2024visual,o1}.
Various LLM reasoning techniques are proposed to improve the LLM performance.
\citet{chen2024language} formulates reasoning as sampling from a latent distribution and optimizing it via variational approaches.
\citet{ho2022large} utilizes LLM as reasoning teachers, improving the reasoning abilities of smaller models through knowledge distillation. 
Among them, Chain-of-Thought (CoT) prompting has emerged as a key technique for improving LLM reasoning by breaking problems into intermediate steps, enabling better performance on multiple tasks~\cite{2022-CoT,lyu2023faithful,li2023making,feng2024towards}.
Extensions of CoT include self-consistency, which aggregates multiple reasoning paths to improve robustness~\cite{wang2022self}, and Tree-of-Thoughts, which explores reasoning steps in a tree-like structure for more complex tasks~\cite{2024-ToT}. 
Reflexion introduces iterative refinement, where the model critiques and updates its intermediate steps~\cite{shinn2024reflexion}.

\noindent\textbf{Token Cost of LLM.}
Although the above methods enhance reasoning accuracy, they often increase token usages, posing challenges to efficiency~\cite{wang2024reasoning,chiang2024over,bhargava2023s}.
Consequently, it is important to mitigate token consumption while maintaining the model performance.
To address this issue, \citet{li2021addressing} introduces a multi-hop processing technique designed to filter out irrelevant reasoning. While effective, this approach is limited to traditional neural networks, such as PALM~\cite{bi2020palm}, and lacks adaptability to large language models (LLMs).
\revise{
Speculative decoding~\cite{leviathan2023fast} aims to accelerate decoding by generating drafts using smaller models and verifying them with larger models, which is over-dependent on the alternative small approximation model.
LLM routing~\cite{dinghybrid} queries to different LLMs based on quality-cost trade-offs, but it cannot reduce the token usage on the specific LLM for a given query.
}
\citet{zheng2024response} aims to improve LLM inference speed by predicting response lengths and applying a scheduling algorithm to enhance efficiency.
However, it is constrained to scheduling level, and it does not reduce the actual token costs. 
\citet{hao2024training} reduces token usage by substituting decoded text tokens with continuous latent tokens. However, its application is currently restricted to small-scale, early language models like GPT-2~\cite{radford2019language}. Additionally, it significantly impacts reasoning accuracy, resulting in over a 20\% relative accuracy reduction on benchmarks such as GSM8K~\cite{2021-GSM8K}.

%% file: sections/observation.tex
\vspace{-0.2cm}
\section{Token Redundancy in LLM Reasoning}
\label{sec:token_redundancy}
\vspace{-0.2cm}
\noindent\textbf{Token Budget.}
Previous research \cite{2024-CCoT} demonstrates that LLM has the potential to follow a length constraint in the prompt.
\autoref{tab:prompt_construction} shows the difference between the vanilla CoT and the CoT with token budget.
For instance, by including a token budget (50 tokens) within the prompt, as illustrated in~\autoref{fig:case_budget_CoT}, the LLM adjusts the length of its output (86 output tokens), trying to align with the specified budget.
This indicates that LLMs have a certain capability in following prompts with an explicit token budget.

\begin{table}[]
    \centering
    \footnotesize
    \tabcolsep=5pt
        \caption{Illustrations of the vanilla CoT prompt and the token-budget-aware prompt.}
\vspace{-0.2cm}  
\begin{tabular}{cm{4cm} p{4cm}}
\toprule
\multicolumn{1}{c}{Prompt method} & \multicolumn{1}{c}{Content} \\
\cmidrule(lr){1-2}
Vanilla CoT & Let's think step by step: \\
\cmidrule(lr){1-2}
CoT with Token Budget & Let's think step by step and use less than \textcolor{red}{budget} tokens: \\
\cmidrule(lr){1-2}
Example & Let's think step by step and use less than \textcolor{red}{50} tokens: \\
\bottomrule
\end{tabular}
\label{tab:prompt_construction}
\vspace{-0.5cm}
\end{table}

\noindent\textbf{Token Redundancy Phenomenon.}
We find that providing a reasonable token budget can significantly reduce the token cost during reasoning.
As shown in \autoref{fig:case_budget_CoT}, including a token budget in the instructions reduces the token cost in the chain-of-thought (CoT) process by several times, but the LLM still gets the correct answer. 
Our results in \autoref{fig:motivation_elastic_observation} and \autoref{tab:main_eval} also confirm there are a large number of redundant tokens in the reasoning process of the state-of-the-art LLMs.

\noindent\textbf{Causes of Token Redundancy in LLM Reasoning.} A possible explanation for this token redundancy is that during the post-training phase, such as the RLHF process~\cite{ouyang2022training}, annotators might favor more detailed responses from LLMs, marking them as preferred. As a result, the model learns to associate longer, more detailed responses with alignment to human preferences and tends to produce such outputs during reasoning. However, in many scenarios, we primarily need LLMs to provide the correct answer and make accurate decisions, rather than elaborate extensively with detailed explanations. This motivates the need to eliminate redundant tokens in the LLM reasoning process in many cases.

\vspace{-0.1cm}
\section{Searching Optimal Token Budget}
\label{sec:observation}
\vspace{-0.1cm}

As demonstrated in \autoref{fig:case_study_cot}, different token budgets have different effects. Therefore, it is natural to investigate the following question: ``\emph{How to search the optimal token budget for a specific question and a particular LLM?}''

\input{algorithm/binary_search}

\begin{figure*}[!thbp]
\centering
     \subfloat[GPT-4o-mini budget search.]
    {
        \includegraphics[width=0.23\linewidth]{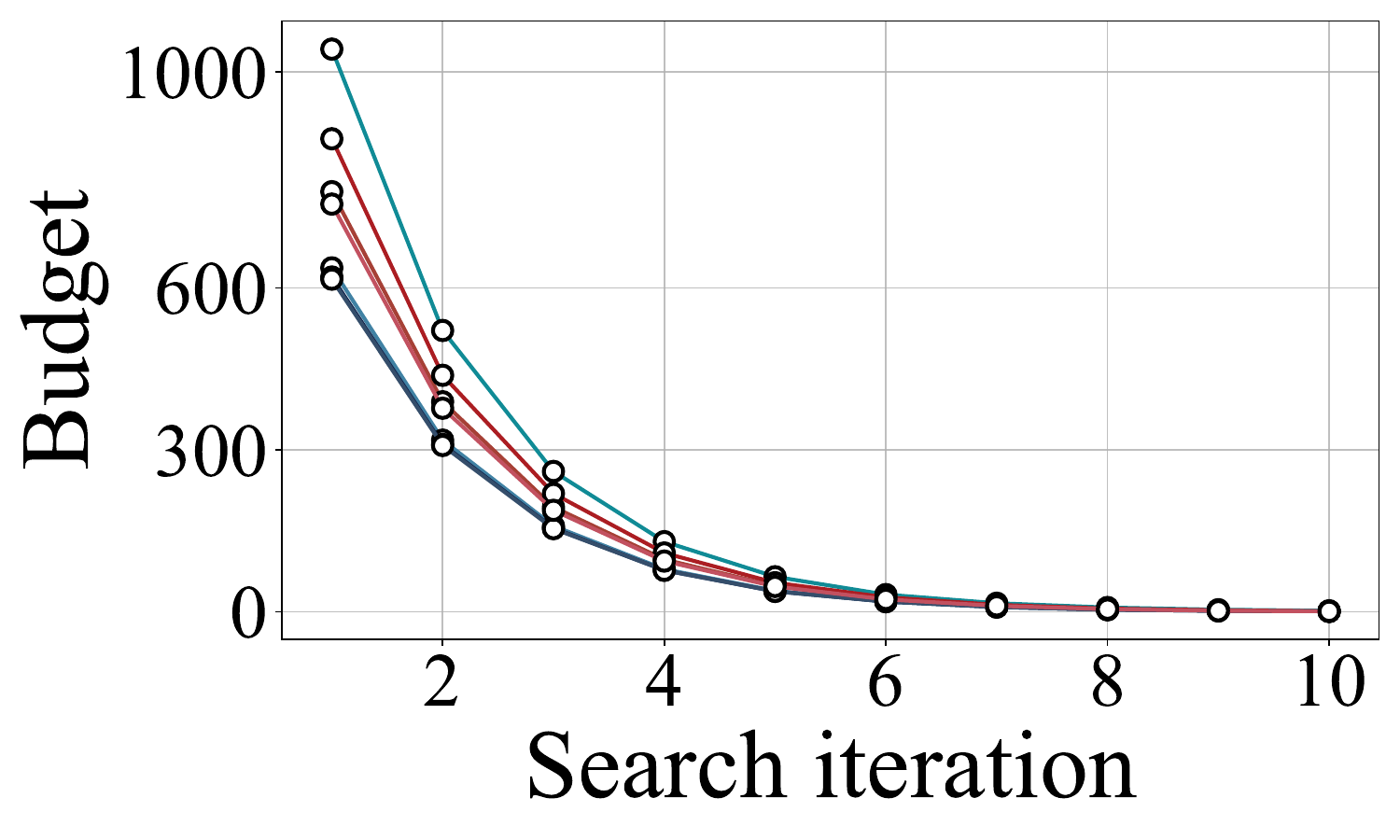}
        \label{fig:gpt-4o-mini-search-budget}
    }
    \subfloat[GPT-4o-mini token cost.]
    {
        \includegraphics[width=0.23\linewidth]{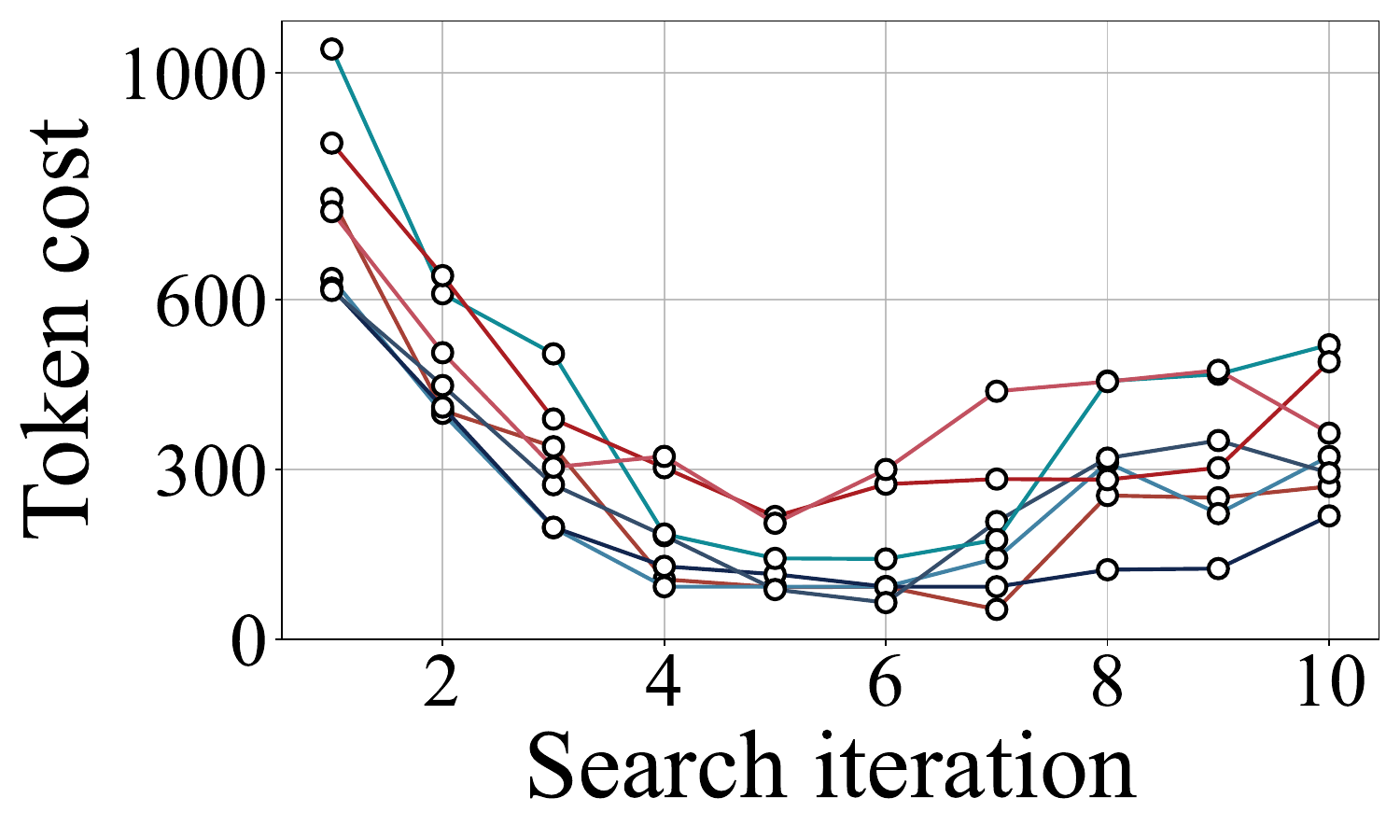}
        \label{fig:gpt4omini_token_cost}
    }
    \subfloat[Yi-lightning budget search.]
    {
        \includegraphics[width=0.23\linewidth]{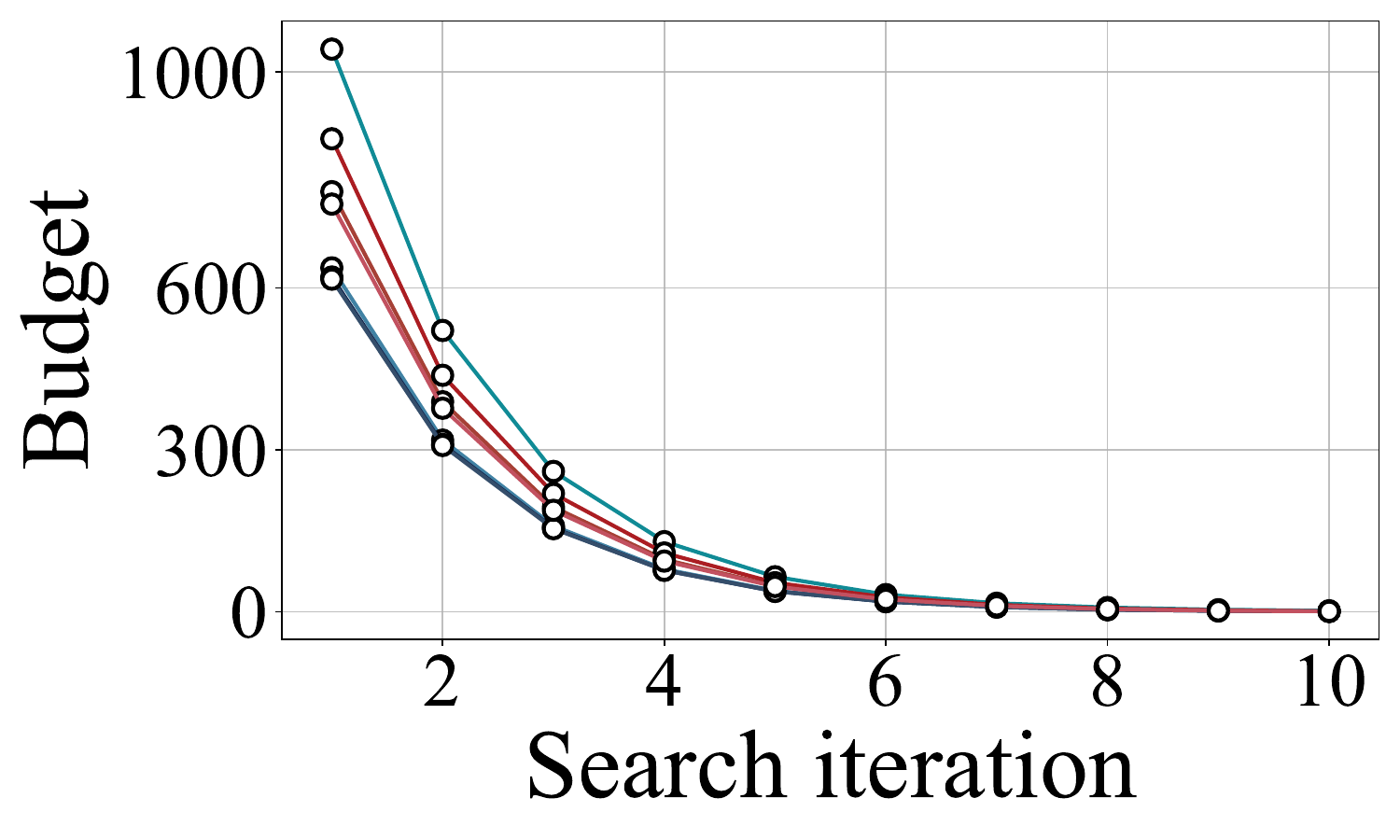}
        \label{fig:yi-lightning-search-budget}
    }
    \subfloat[Yi-lightning token cost.]
    {
        \includegraphics[width=0.23\linewidth]{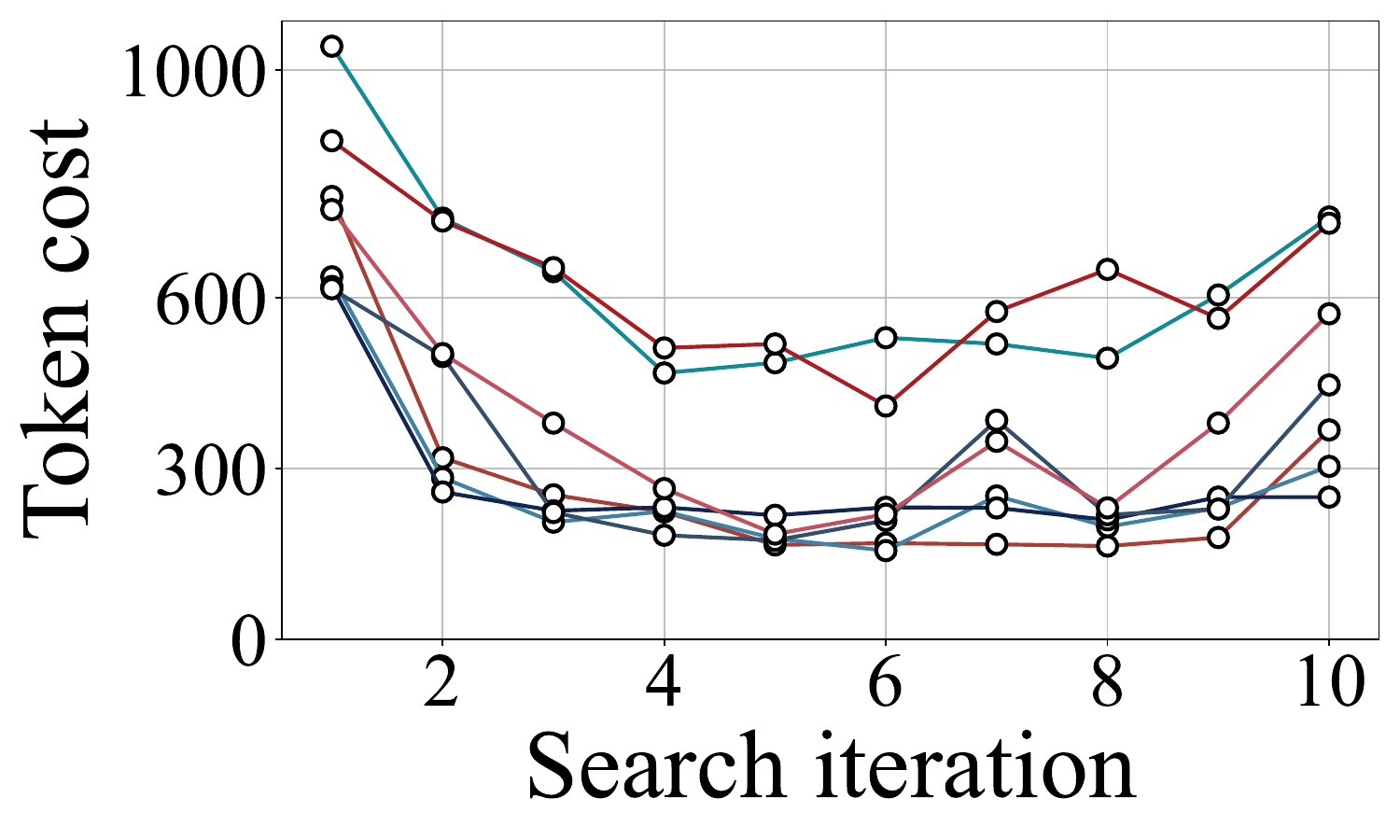}
        \label{fig:yilightning_token_cost}
    }
    \caption{Token elasticity phenomenon. The x-axis denotes the budget search iteration. The y-axis denotes the searched budget (\autoref{fig:gpt-4o-mini-search-budget} and \autoref{fig:yi-lightning-search-budget}) or the real token costs for each searched budget (\autoref{fig:gpt4omini_token_cost} and \autoref{fig:yilightning_token_cost}).
    Different colors denote different samples randomly selected from MathBench-College~\cite{2024-mathbench}.
    The token cost is significantly lower in a reasonable token budget range. When the token budget is smaller than the reasonable range, the token cost gradually increases.}
    \label{fig:motivation_elastic_observation}
    \vspace{-0.3cm}
\end{figure*}

\noindent\textbf{Vanilla Method for Optimal Budget Search.}
An intuitive method is finding the minimal needed tokens as the budget, ensuring that the LLM can still produce correct and accurate responses within this constraint.
\revise{
The goal of the search algorithm is not to directly determine task difficulty, but to identify the minimum token budget under which the model can still produce a correct answer. Specifically, the algorithm conducts a binary search over different token budgets and selects the shortest budget that maintains correctness. This approximates the minimum reasoning length required for the model to solve the given problem.
}\revise{
To find the minimal token budget, we first propose an ``implicit monotonicity assumption'' that when the model outputs a wrong prediction at a budget value, it always predicts incorrectly when under this budget value, and when the model outputs a correct prediction at a budget value, it always predicts correctly when above this budget value.
To empirically assess the validity of this assumption, we conduct an additional analysis. Specifically, we define a sample as monotonic if all predictions above the optimal budget are correct, and all predictions below it are incorrect.
We randomly sample from the GSM8K dataset as test data and find that 90.91\% of the samples satisfy this monotonicity condition. This suggests that while the assumption may not strictly hold in every instance, it is a reasonable and effective approximation in practice for guiding budget search.
An intuitive monotonic sample is illustrated as \autoref{tab:intuitive_monotonicity}.
Based on the ``implicit monotonicity assumption'', we further design a binary search-based minimal budget search algorithm detailed in \autoref{alg:binary_search}.
}
\begin{table}[!t]
    \centering
    \small
    \tabcolsep=5pt
    \renewcommand{\arraystretch}{1.2}
    \caption{An intuitive monotonic example. $\beta^*$ is the searched optimal budget. The budget row displays scaled budgets ranging from $2^{-2}$ to $2^{2} \cdot \beta^*$.}
\vspace{-0.2cm}
\begin{tabular}{lccccc}
\toprule
\textbf{Budget($*\beta^*$)} & $2^{-2}$ & $2^{-1}$ & $1$ & $2^{1}$ & $2^{2}$ \\ 
\midrule
\textbf{Prediction} & False & False & True & True & True \\ \bottomrule
\end{tabular}
\label{tab:intuitive_monotonicity}
 \vspace{-0.5cm}
\end{table}

Before initiating the search process, we first apply the vanilla CoT to generate an answer for each question, as illustrated in \autoref{fig:case_vanilla_CoT}. The number of tokens in the resulting answer is then calculated and designated as the right boundary for search, denoted by $right$.
The function \texttt{isFeasible} is used to determine the feasibility of a budget. A budget is considered feasible here if the CoT prompt with that budget preserves the correctness of the answer.
\autoref{alg:binary_search} showcases the details.
Given the feasibility function, large language model $\mathcal{M}$, question $\vx$ and label $y$ as the input, \autoref{alg:binary_search} first calculates the right boundary of search (line 2).
With $0$ as the left boundary, the current possible budget $\beta$ is computed as the midpoint of $0$ and $right$ (line 3).
We use $\beta_0$ to record the previously searched budget (line 4).
While the current $\beta$ is feasible, the algorithm updates $\beta$ by recalculating the midpoint (line 7) and adjusts the search bounds accordingly to narrow the range (line 9). 
Once the loop ends, the final budget $\beta$ is returned as the searched result (line 12). 
\autoref{alg:binary_search} is designed to find the minimal budget efficiently.
However, we observe that the minimal budget required to produce a correct answer is not necessarily the optimal budget. When the budget is unreasonably small, the actual token cost often exceeds that of cases where a larger budget is used. 
\revise{
We also further formalize the optimal budget search process detailed in \autoref{subsec:formalizing_budget_search}.
}
\begin{figure}[t]
    \centering
  \includegraphics[width=0.88\columnwidth]{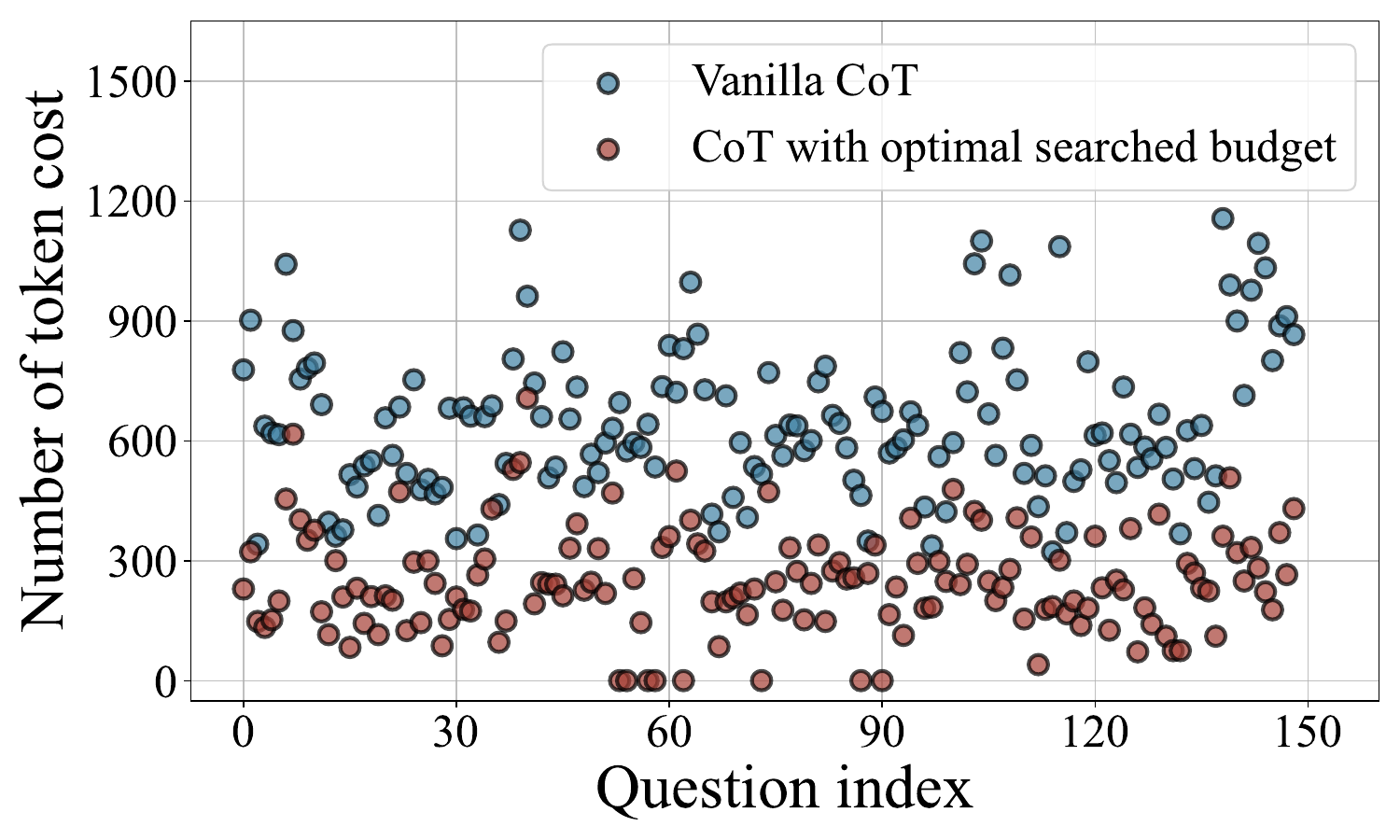}
  \vspace{-0.3cm}
  \caption{The effects of optimal searched budget. CoT, with our optimal searched budget, reduces the token costs significantly without influencing the accuracy. 
  We conduct it on MathBench-College~\cite{2024-mathbench}.}
  \label{fig:token_costs_optimal_budget}
  \vspace{-0.4cm}
\end{figure}

\noindent\textbf{Observation of Token Elasticity.}
During our minimal budget search process, we observe a ``\textit{token elasticity}'' phenomenon as we approach the minimal budget. 
Specifically, as \autoref{alg:binary_search} progresses, we aim to identify the minimal budget that still ensures the answer's correctness. 
However, we find that if the budget is reduced beyond a certain range, the token cost increases, indicating that further reductions in the budget lead to increasing token consumption.
\autoref{fig:motivation_elastic_observation} showcases the evidence.
The x-axis represents the iterations of the budget binary search, with the budget values decreasing progressively.
The y-axis in \autoref{fig:gpt4omini_token_cost} and \autoref{fig:yilightning_token_cost} show the corresponding token costs at each budget search iteration.
\revise{
When the searched budget drops below a reasonable range, the token cost increases.
This happens because the model, unable to meet the tight constraint, ignores it and reverts to longer reasoning.
In other words, the model effectively ``gives up'' on complying with the instruction, resulting in longer outputs and redundant token costs. This explains the non-monotonicity observed in Figure 2: token usage initially decreases as the budget tightens but eventually rebounds when the budget becomes too small to be feasible. We will make this intuition clearer in the revised version.
}
\autoref{fig:one-budget} also shows an example. As observed, when a small token budget (e.g., 10 tokens) is used, the real token cost is significantly higher compared to scenarios where a reasonable token budget is allocated (i.e., \autoref{fig:case_budget_CoT}).

\input{algorithm/vanilla_feasibility}

\noindent\textbf{Token Elasticity based Optimal Budget Search.}
The token elasticity observation shows that while a minimal budget may keep the correctness of the answer, it does not necessarily minimize the token cost.
\autoref{fig:one-budget} and \autoref{fig:case_budget_CoT} illustrate an intuitive example.
To address this, we enhance \autoref{alg:binary_search} by incorporating a greedy search strategy aimed at finding the optimal budget that simultaneously minimizes token cost and preserves answer correctness.
Specifically, we introduce an additional constraint to the \texttt{isFeasible} condition.
Beyond ensuring correctness, the updated budget must result in a lower token cost compared to the previously searched budget. 
\autoref{alg:greedy_feasibility} outlines the feasibility function employed during the search process. Initially, the actual token cost is computed for both the current and previously evaluated budgets (line 2). Next, feasibility is assessed based on two criteria: the answer correctness and greedy token reduction (line 3). The search process is terminated if either condition fails.

\revise{
\noindent\textbf{Overhead of Budget Search.}
For the overhead, note that the budget search mainly serves to reveal token compression potential, motivating TALE's design.
For TALE-PT, the budget search is used once offline to generate training targets for post-training. Since this is a one-time pre-processing step, it does not incur additional costs during actual model deployment.
For TALE-EP, we clarify that the budget search is not needed at all. TALE-EP relies on a lightweight, zero-shot budget estimator that directly predicts a reasonable token budget without any iterative search, making it training-free and highly efficient at inference time.
To quantify the budget search overhead for TALE-PT data generation, we measured the total time needed to run the search algorithm and produce the optimal budget dataset. On GSM8K (7473 samples), this process takes approximately 354 minutes on an A100 GPU, which we consider acceptable given that it is a one-time offline cost for training.
}

%% file: algorithm/binary_search.tex
\begin{algorithm}[!t]
    \caption{Budget Search}
    \label{alg:binary_search}
    \begin{algorithmic}[1]
        \Require feasibility checking function \texttt{isFeasible}, a large language model $\mathcal{M}$, a given question $\vx$ and the ground truth label $y$
        \Ensure searched budget $\beta$
    \Function {Search}{\texttt{isFeasible},$\mathcal{M},\vx,y$}
        \State $right \gets$ the actual token costs of $\mathcal{M}$ with vanilla CoT prompt on $\vx$
        \State $\beta \gets \lfloor (0 + right) / 2 \rfloor$
        \State $\beta_0 \gets right$
        \While{\texttt{True}}
        \If{\texttt{isFeasible}$(\mathcal{M}, \vx, y, \beta_0, \beta)$}
            \indent\indent\Comment{Update the searched budget}
            \vfill
            \State $\beta \gets \lfloor (0 + right) / 2 \rfloor$ 
            \vfill 
            \indent\indent\Comment{Record previous searched budget}
            \State $\beta_0 \gets right$
            \vfill 
            \indent\indent\Comment{Update the search range}
            \State $right \gets \beta$ 
        \Else
            \State \textbf{break}
        \EndIf
        \EndWhile
    \State \Return $\beta$ 
    \EndFunction
  \end{algorithmic}
  
\end{algorithm}

%% file: algorithm/vanilla_feasibility.tex
\begin{algorithm}[!t]
\caption{Greedy Feasibility Function}
\label{alg:greedy_feasibility}
\begin{algorithmic}[1]
\Require a large language model $\mathcal{M}$, a question $\vx$ and the ground truth label $y$, previous and current budget: $\beta_0, \beta$
\Ensure \texttt{True} if the budget satisfies the requirements, \texttt{False} otherwise
\Function {\texttt{isFeasible}}{$\mathcal{M}, \vx, y, \beta_0,\beta$}
    \State $t, t_0 \gets$ gets the actual token costs under budgets of $\beta$ and $\beta_0$
    \If{$\mathcal{M}(\vx, \beta) == y$ \textbf{and} $t < t_0$}
    \State\Return{\texttt{True}}
    \EndIf
    \State \Return{\texttt{False}}
\EndFunction
\end{algorithmic}
\end{algorithm}

%% file: sections/methodology.tex
\vspace{-0.1cm}
\section{Methodology}
\label{sec:meth}
\vspace{-0.1cm}

\subsection{Overview}
\begin{figure}[!t]
    \centering
    \includegraphics[width=0.95\linewidth]{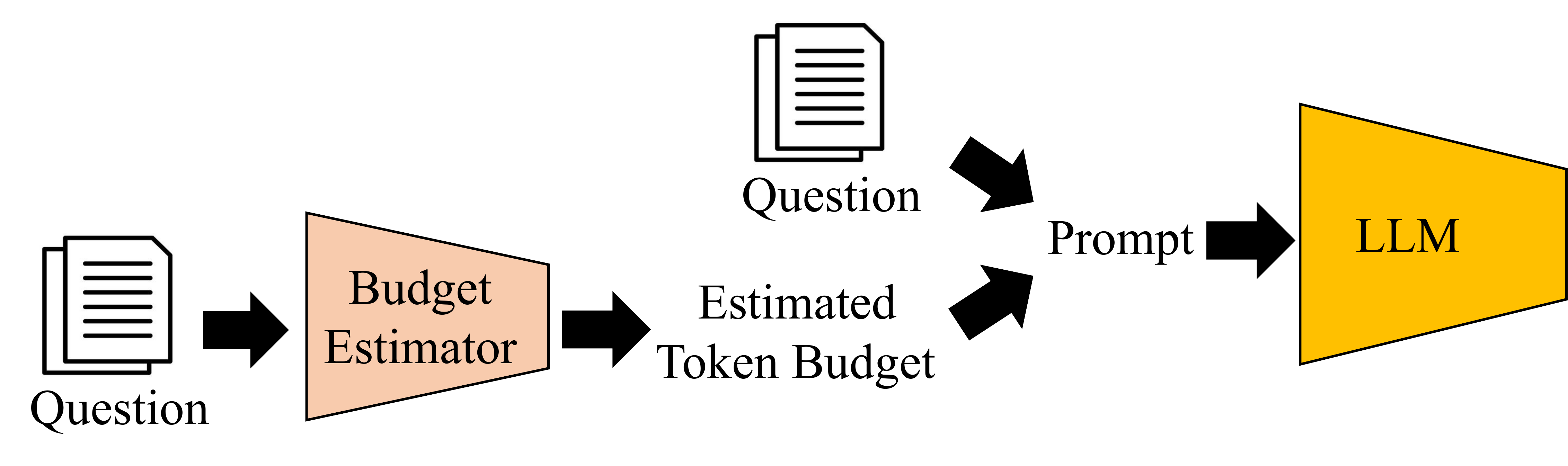}
    \vspace{-0.3cm}
    \caption{The workflow of \ours{}-EP.
    Given a question, \ours{}-EP first estimates the token budget using a budget estimator. It then crafts a token-budget-aware prompt by combining the question with the estimated budget. Finally, the prompt is input to the LLM to generate the answer as the final output. By default, we use the reasoning LLM itself with zero-shot estimation prompt as the budget estimator.
    }
    \label{fig:framework}
    \vspace{-0.1cm}
\end{figure}

\begin{figure}[]
    \centering
    \includegraphics[width=0.89\linewidth]{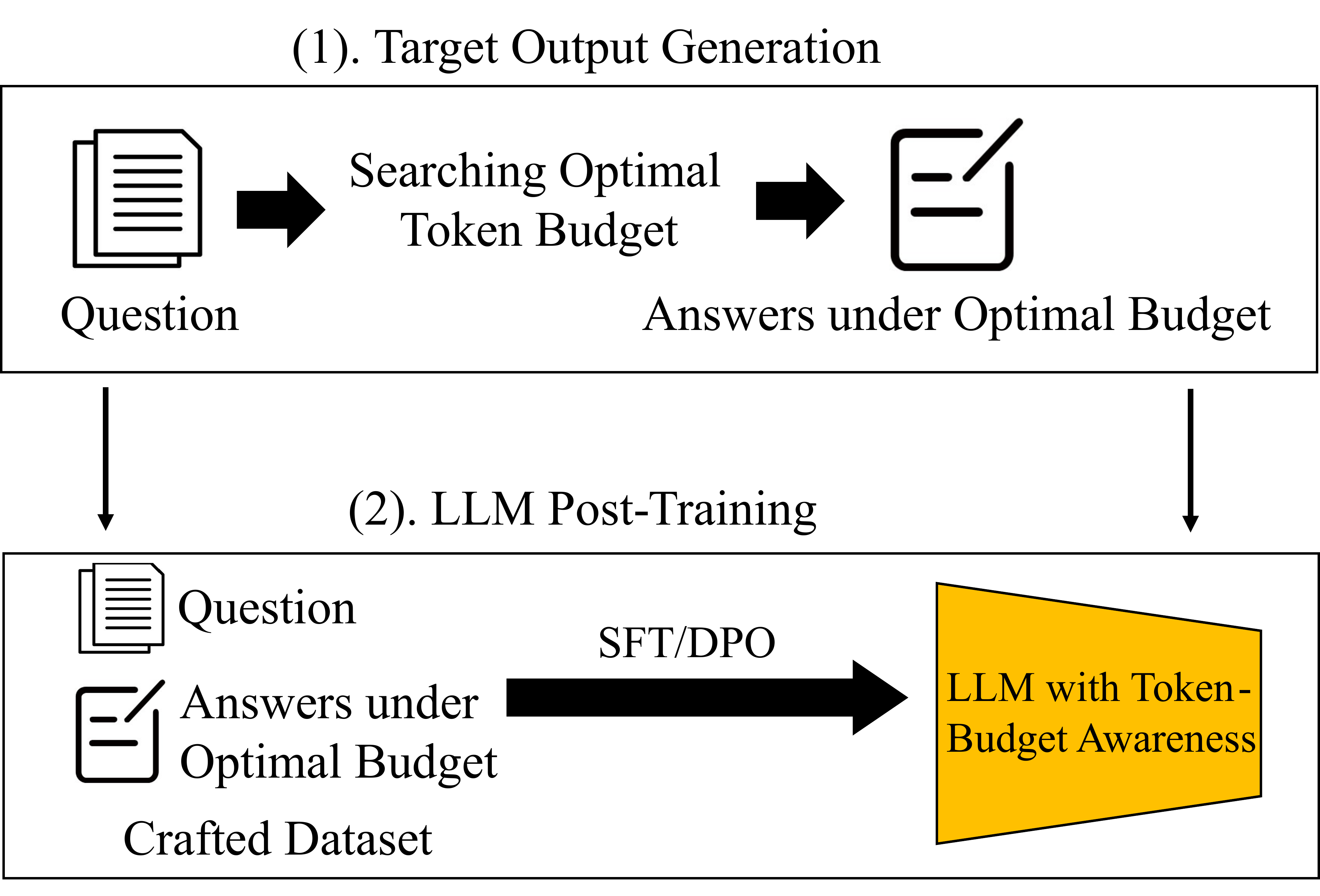}
    \vspace{-0.2cm}
    \caption{The workflow of \ours{}-PT.
    Given a set of questions, \ours{}-PT first generates target outputs in phase (1), searching the answers under optimal budget. In phase (2), \ours{}-PT uses the searched target outputs to craft a specialized dataset, then post-train the LLM (via SFT/DPO) to internalize token-budget awareness.}
    \label{fig:post_training}
    \vspace{-0.4cm}
\end{figure}

\vspace{-0.1cm}
Based on the above analysis, we designed our method \sys for token-budget-aware reasoning in LLMs.
Two solutions, i.e., estimation\&prompting (\ours{}-EP, see \autoref{fig:framework}) and post-training (\ours{}-PT, see \autoref{fig:post_training}), are proposed.

\begin{figure}[!t]
    \centering
    \includegraphics[width=0.95\linewidth]{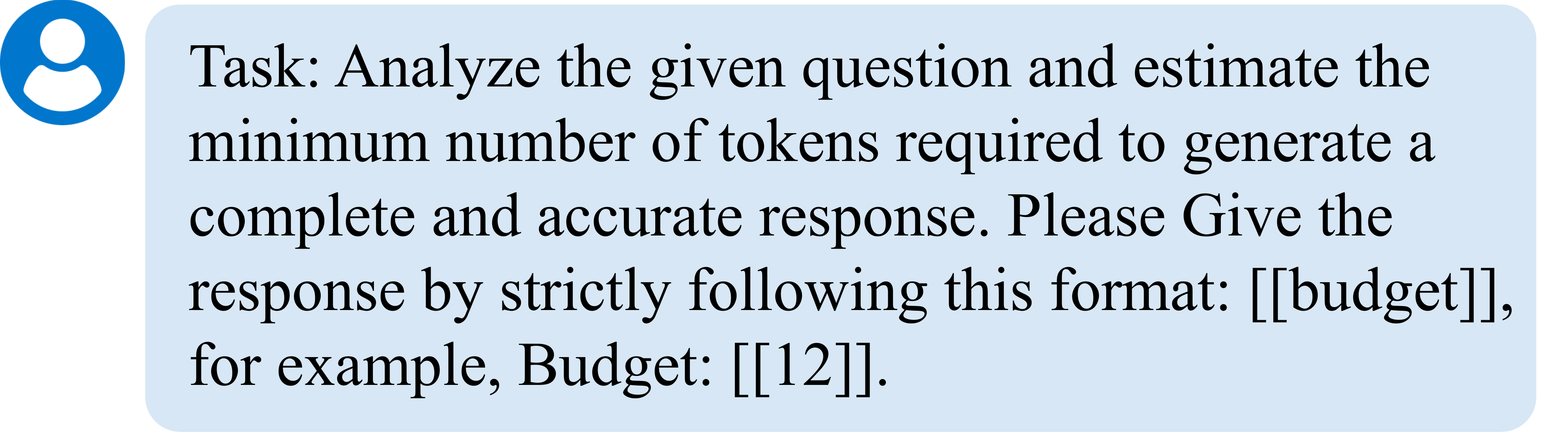}
    \vspace{-2mm}
    \caption{The prompt for zero-shot budget estimation.}
    \label{fig:zero_prompt}
    \vspace{-4mm}
\end{figure}

\vspace{-0.2cm}
\subsection{Estimation and Prompting (\ours{}-EP)}
\label{subsec:tale_estimation_prompting}
\vspace{-0.1cm}
Our observations on token elasticity (\autoref{sec:observation}) indicate that only a well-chosen budget within a reasonable range can effectively minimize token costs while preserving LLM performance. The optimal budget, found using \autoref{alg:binary_search} and \autoref{alg:greedy_feasibility}, lies within this range and achieves a satisfying trade-off between efficiency and performance.
Building on this insight, we introduce a token budget aware reasoning method by zero-shot-based token budget estimation and prompting the reasoning LLM. \ours{}-EP leverages the reasoning capabilities of the LLM as an estimator. \autoref{fig:framework} provides an overview of \ours{}-EP’s workflow.
The goal of \ours{}-EP is to construct a token-budget-aware prompt that maintains performance comparable to vanilla CoT while reducing token costs. To achieve this balance, \ours{}-EP follows a two-phase approach: budget estimation and prompt construction.
Given a question, \ours{}-EP first estimates a reasonable token budget that closely aligns with the optimal searched budget. 
By default, we use
the reasoning LLM itself with a zero-shot estimation prompt as the budget estimator.
\autoref{fig:zero_prompt} demonstrates the budget estimation prompt, \revise{which will guide the model to evaluate the question as a whole.}.
Using this estimate, it then crafts a token-budget-aware prompt and feeds it into the LLM to generate the final answer. 
\autoref{fig:intuitive_example_workflow_ep} illustrates this process with a concrete example.
The key intuition behind \ours{}-EP is inspired by human-like thinking. When solving a mathematical problem, a person may take time to compute the exact answer but can quickly estimate the effort required to solve it. For instance, when comparing a primary school arithmetic question to a college-level calculus problem, one may not immediately provide the solutions but can easily infer that the former takes only seconds while the latter requires significantly more time.
\autoref{subsec:rq2_budget_estimation} evaluates the effectiveness of our budget estimation approach, demonstrating that the budgets estimated by advanced LLMs (e.g., GPT-4o-mini) are generally close to the optimal searched budget and deliver competitive performance.

\subsection{\ours{} Post-Training (\ours{}-PT)}
\label{subsec:tale_post_training}
Another approach for obtaining an LLM with token-budget awareness is post-training it to incorporate this awareness into its inference process, enabling it to generate more token-efficient reasoning responses.
Specifically, we post-train the LLM $\mathcal{M}_{\theta}$ to produce answers that adhere to the token budget.
This process is divided into two key stages: target output generation and LLM post-training.

\noindent\textbf{Target Output Generation.}
In the target output generation stage, we craft the target output $y_i$ by prompting $\mathcal{M}_{\theta}$ with a Chain-of-Thought (CoT) prompt that incorporates our searched optimal token budget.
The prompt is formatted as follows:
\begin{center}
    \texttt{``Let's think step by step and use less than $\beta_{i}^*$ tokens:''}
\end{center}
where $\beta_{i}^*$ is the searched optimal budget for the given question $\vx_i$ (see search process in \autoref{alg:binary_search} and \autoref{alg:greedy_feasibility}).
\autoref{fig:case_budget_CoT} illustrates an example.
The resulting LLM output, constrained by the token budget specified in the prompt, is taken as the crafted target output $y_i$. This target output not only leads to the correct answer but also has minimal actual output token cost among our token elasticity-based search process, as described in \autoref{sec:observation}.
In the LLM post-training stage, we train the LLM $\mathcal{M}_{\theta}$ using the crafted target outputs from the first stage. We introduce two ways to conduct the token-budget awareness internalization during post-training, i.e., SFT-based and DPO-based method.
Details of the hype parameters are in \autoref{subsec:details_implementation}.

\noindent\textbf{SFT-based Internalization.}
To inject token-budget awareness into $\mathcal{M}_{\theta}$, we perform supervised fine-tuning with these target outputs.
We post-train $\mathcal{M}_{\theta}$ to generate token-efficient outputs by minimizing the cross-entropy loss between the model’s predictions and the target outputs. Given an input $\vx$ and a target output $y$ from the first stage (which reflects token-budget awareness), the cross-entropy loss is defined as:
\begin{equation*}
    \mathcal{L}_{\text{CE}}(\theta) = - \frac{1}{N} \sum_{i=1}^{N} \sum_{t=1}^{T_i} \log \mathbb{P}(y_{i,t} | y_{i,<t}, \vx_i),
\end{equation*}
where $T_i$ means the length of the target sequence $y_i$ for the $i$-th training example, $y_{i,t}$ the target token at position $t$ of $y_i$, $y_{i,<t}$ means the sequence of tokens preceding the current token $y_{i,t}$, representing the context up to time step $t$ for the $i$-th sample.
$\mathbb{P}(y_{i,t} | y_{i,<t}, x_i)$ represents the conditional probability predicted by the model $\mathcal{M}_{\theta}$ for the token $y_{i,t}$, given the input $\vx_i$ and the preceding tokens $y_{i,<t}$.
The loss is based on the next token prediction.
The goal is to adjust the model parameters $\theta$ such that it produces concise and accurate responses that adhere to the token budget constraint. This is achieved through gradient descent, forcing the model to internalize the compact reasoning patterns from the token-efficient target outputs.

\input{tables/main_eval}

\input{tables/generalization}

\input{tables/internalization}

\noindent\textbf{DPO-based Internalization.}
Another way to incentivize $\mathcal{M}_{\theta}$ to learn the token-budget preference is applying the DPO algorithm~\cite{2023-DPO} to post-train the model.
DPO directly refines the policy through a classification objective, aligning the model's behavior with the desired preferences. 
The goal of DPO here is to refine $\mathcal{M}_{\theta}$ so it can accurately solve a given problem $\vx$ while adhering to an internalized token budget.
We use the target outputs $y_i$ from the searched optimal budget as positive samples, while outputs $y_i'$ generated with the vanilla CoT prompt serve as negative samples. These positive-negative pairs are then used to create the pairwise preference data for DPO training.
Given the crafted dataset $\mathcal{D}$ = $\{(\vx_i, y_i, y_i')\}_{i=1}^{N}$,
the objective is to maximize the likelihood that the model ranks the positive samples higher than the negative ones.
Formally, we aim to optimize the following objective:
\begin{equation*}
\begin{aligned}
\mathcal{L}_{\mathrm{DPO}}(\theta)
&= -\frac{1}{N} \sum_{i=1}^{N} \log P_{\theta}(y_i \succ y_i'), \quad where\\
P_{\theta}(y_i \succ y_i')&=
    \frac{\exp\bigl(s(y_i, \mathbf{x}_i)\bigr)}
         {\exp\bigl(s(y_i, \mathbf{x}_i)\bigr)
          + \exp\bigl(s(y_i', \mathbf{x}_i)\bigr)}.
\end{aligned}
\end{equation*}

$P_{\theta}(y_i\succ y_i')$ is the preference function. Here, $s(y_i, \vx_i)$ is defined as $\sum_{t=1}^{T_i}\log \mathbb{P}(y_{i,t} | y_{i,<t}, \vx_i)$, and it represents the log-probability of the model generating $y_i$ for input $\vx_i$, which serves as the preference score assigned to $y_i$.
This score measures how strongly the model favors that output.
The objective ensures that the model prioritizes concise and token-efficient outputs while maintaining high-quality reasoning and correctness.
During training, the LLM is encouraged to internalize the token budget constraint and adopt a more compact reasoning process guided by the target outputs generated in the first stage.
This two-stage process effectively trains the LLM to produce concise yet accurate responses, striking a balance between reasoning quality and token efficiency during inference.
More details are in \autoref{subsec:details_implementation}.

%% file: tables/main_eval.tex
\begin{table*}[th]
    \centering
    \scriptsize
    \tabcolsep=3pt
    \caption{Comparison of \ours{}-EP (estimation and prompting) and other prompt engineering methods.
    ``Directly Answering'' means prompting LLM without any reasoning process.
    ``Vanilla CoT'' means the vanilla CoT prompting without budget.
    The model used in our evaluation is GPT-4o-mini~\cite{gpt4o-mini2024}.
    Observe that \ours{}-EP achieves an average accuracy (ACC) of 80.22\%, with an average output token cost of 138.53 and an average expense of 118.46.
    \ours{}-EP reduces output token costs by 67\%, lowers expenses by 59\%, and maintains competitive performance compared to the vanilla CoT approach.
    ACC $\uparrow$, Output Tokens $\downarrow$, Expense ($10^{-5}\$$ / sample) $\downarrow$.
    }
    \vspace{-0.2cm}
   \begin{tabular}{lccccccccc}
\toprule
\multirow{2}{*}{Dataset} & \multicolumn{3}{c}{Directly Answering} & \multicolumn{3}{c}{Vanilla CoT} & \multicolumn{3}{c}{\ours{}-EP} \\
\cmidrule(lr){2-4} \cmidrule(lr){5-7} \cmidrule(lr){8-10}
 & ACC $\uparrow$ & Output Tokens $\downarrow$ & Expense $\downarrow$ & ACC $\uparrow$ & Output Tokens $\downarrow$ & Expense $\downarrow$ & ACC $\uparrow$ & Output Tokens $\downarrow$ & Expense $\downarrow$ \\
 \cmidrule(lr){1-10}
GSM8K & 28.29\% & 12.46 & 39.43 & 81.35\% & 318.10 & 541.09 & 84.46\% & 77.26 & 279.84 \\
GSM8K-Zero & 97.21\% & 18.85 & 91.69 & 99.50\% & 252.96 & 886.79 & 98.72\% & 22.67 & 276.12 \\
MathBench-Arithmetic & 59.67\% & 41.10 & 9.78 & 75.00\% & 313.51 & 78.58 & 73.67\% & 39.60 & 18.62\\
MathBench-Middle & 33.33\% & 5.00 &  3.58 & 84.67\% & 553.93 & 68.22 & 79.33\% & 238.14 & 42.95\\
MathBench-High & 51.33\% & 5.00 & 4.07 & 84.00\% & 653.24 & 82.44 & 80.00\% & 254.82 & 47.61\\
MathBench-College & 44.00\% & 5.00 & 3.68 & 78.00\% & 675.78 & 81.56 & 70.00\% & 259.85 & 45.60\\
\cmidrule(lr){1-10}
Average & 52.31\% & 14.57 & 25.37 & 83.75\% & 461.25 & 289.78 & 81.03\% & 148.72 & 118.46 \\
\bottomrule
\end{tabular}
    \label{tab:main_eval}
    \vspace{-0.1cm}
\end{table*}

%% file: tables/generalization.tex
\begin{table*}[th]
    \centering
    \scriptsize
    \tabcolsep=3pt
    \caption{The generalization of \ours{}-EP (estimation and prompting) across different LLMs. Yi-lightning~\cite{2024-yi-lightning}, GPT-4o-mini~\cite{gpt4o-mini2024}, GPT-4o~\cite{gpt4o-2024} and o3-mini~\cite{o3-mini} are taken into consideration. We conduct following evaluations on the MathBench-College dataset. ACC $\uparrow$, Output Tokens $\downarrow$, Expense ($10^{-5}\$$ / sample) $\downarrow$.}
    \vspace{-0.2cm}
   \begin{tabular}{lccccccccc}
\toprule
\multirow{2}{*}{LLM} & \multicolumn{3}{c}{Directly Answering} & \multicolumn{3}{c}{Vanilla CoT} & \multicolumn{3}{c}{\ours{}-EP} \\
\cmidrule(lr){2-4} \cmidrule(lr){5-7} \cmidrule(lr){8-10}
 & ACC $\uparrow$ & Output Tokens $\downarrow$ & Expense $\downarrow$ & ACC $\uparrow$ & Output Tokens $\downarrow$ & Expense $\downarrow$ & ACC $\uparrow$ & Output Tokens $\downarrow$ & Expense $\downarrow$ \\
 \cmidrule(lr){1-10}
Yi-lightning & 66.67\% & 80.01 & 3.09 & 79.33\% & 998.10 & 21.55 & 76.67\% & 373.52 & 17.25\\
GPT-4o-mini & 44.00\% & 5.00 & 3.68 & 78.00\% & 675.78 & 81.56 & 70.00\% & 259.85 & 45.60 \\
GPT-4o & 57.33\% & 5.00 & 61.34 & 84.00\% & 602.29 & 1359.42 & 80.00\% & 181.61 & 759.95 \\
o3-mini & 96.00\% & 601.51 & 336.69 & 97.33\% & 1163.55 & 638.46 &  96.66\%& 677.65 & 385.12 \\
\bottomrule
\end{tabular}
    \label{tab:generalization}
    \vspace{-0.4cm}
\end{table*}

%% file: tables/internalization.tex
\begin{table*}[th]
    \centering
    \scriptsize
    \tabcolsep=3pt
    \caption{Comparison of \ours{}-PT (post-training to internalize token-budget awareness) and other prompt engineering methods.
    Two different post-training methods, SFT and DPO, are taken into consideration.
    }
    \vspace{-0.2cm}
   \begin{tabular}{lccccccccc}
   \toprule
 &  & \multicolumn{2}{c}{Directly Answering} & \multicolumn{2}{c}{Vanilla CoT} & \multicolumn{2}{c}{TALE-PT-SFT} & \multicolumn{2}{c}{TALE-PT-DPO} \\
 \cmidrule(lr){3-4} \cmidrule(lr){5-6} \cmidrule(lr){7-8} \cmidrule(lr){9-10}
\multirow{-2}{*}{LLM} & \multirow{-2}{*}{Dataset} & ACC $\uparrow$ & Output Tokens $\downarrow$ & ACC $\uparrow$ & Output Tokens $\downarrow$ & ACC $\uparrow$ & Output Tokens $\downarrow$ & ACC $\uparrow$ & Output Tokens $\downarrow$ \\
\midrule
 & GSM8K & 21.00\% & 38.54 & 77.56\% & 241.51 & 78.57\% & 139.63 & 74.11\% & 149.93 \\
\multirow{-2}{*}{Llama-3.1-8B-Instruct} & GSM8K-Zero & 70.32\% & 13.49 & 65.04\% & 251.08 & 78.43\% & 77.85 & 78.41\% & 113.41 \\
\bottomrule
\end{tabular}
    \label{tab:internalization}
    \vspace{-0.1cm}
\end{table*}

%% file: sections/evaluation.tex
\vspace{-0.1cm}
\section{Evaluation}
\label{sec:eval}
\vspace{-0.1cm}
In this section, we provide the experiment results to evaluate the effectiveness of two versions of \ours{}, \ours{}-EP and \ours{}-PT.
\revise{
The comparisons of the two implementations are detailed in \autoref{subsec:formalizing_budget_search}.
}

\subsection{Experiment Setup}
\label{subsec:experiment_setup}
\noindent\textbf{Datasets.}
To evaluate the LLM performance, three most challenging mathematical datasets are taken into consideration: GSM8K~\cite{2021-GSM8K}, GSM8K-Zero~\cite{chiang2024over}, and MathBench~\cite{2024-mathbench}.
GSM8K-Zero, derived from the GSM8K dataset, specifically targets the analysis of over-reasoning and redundancy in LLM-generated outputs.
In short, GSM8K-Zero is designed so that the answers are embedded within the questions themselves.
LLMs can easily generate correct responses without complicated additional reasoning or redundant calculations.

\noindent\textbf{Models.}
We conduct experiments on five state-of-the-art LLMs (i.e., GPT-4o~\cite{gpt4o-2024}, GPT-4o-mini~\cite{gpt4o-mini2024}, Yi-lightning~\cite{2024-yi-lightning}, o3-mini~\cite{o3-mini}), and Lllama-3.1-8B-Instruct~\cite{dubey2024llama}.

\noindent\textbf{Metrics.} 
The target of \ours{} is to balance the LLM correctness performance and extra redundant token costs.
Specifically, \ours{} seeks to minimize \textit{Number of Output Tokens} while maintaining comparable \textit{Accuracy (Acc)} simultaneously. 

\noindent
\emph{Accuracy (Acc).}
This metric is calculated as the following:
\(Accuracy = \frac{1}{N}\sum_{i=1}^{N}{\mathbb{I}\{\mathcal{M}(\vx_i) = y_i\}}\),
where $(\vx_i, y_i) \in \mathcal{X}$.
$\vx_i$ is the math question from dataset $\mathcal{X}$ and $y_i$ the ground truth answer.
$\mathcal{M}(\cdot)$ returns the answer for a given question.
$\mathbb{I}\{\cdot\}$ represents an indicator function. 
This function evaluates whether the inside given condition holds. Specifically, it returns \textbf{1} if the condition is true and \textbf{0} if the condition is false.
For a better evaluation, we format the LLM output by crafting an elaborate instruction detailed in \autoref{fig:format_prompt}.

\noindent
\emph{Number of Output Tokens.}
We evaluate the token costs by calculating the average output token consumption for each specific task.
The output token costs are measured as follows:
\(\textit{Number of Output Tokens} = \frac{1}{N}\sum_{i=1}^{N}{\mathbb{T}(\mathcal{M}(\vx_i))}\),
where $\vx_i$ represents the given question, and $\mathbb{T}$ is a function that measures the number of tokens.
Intuitively, the more output tokens, the higher the costs incurred by $\mathcal{M}$.
To evaluate costs more precisely, we calculate the average expense per sample. The total token expense includes both input and output tokens used during the query process.

\subsection{Effectiveness of \ours{}-EP}
\label{subsec:rq1_effectiveness}
\autoref{tab:main_eval} compares \ours{}-EP with other prompt engineering methods across seven datasets, evaluating accuracy, output tokens, and expenses. Effective prompts should maximize accuracy while minimizing token usage and cost. 
Direct Answering is the most cost-efficient (14.57 tokens, 25.37 expense) but with low accuracy (52.31\%). Vanilla CoT achieves the highest accuracy (83.75\%) but at a high cost (461.25 tokens, 289.78 expense). \ours{}-EP balances performance and efficiency, achieving 81.03\% accuracy while reducing token usage to 32\% and expenses to 41\% of Vanilla CoT. On GSM8K, it even surpasses Vanilla CoT with 84.46\% accuracy.
Note that expense is not directly proportional to output tokens because it also accounts for input and cached tokens.
\ours{}-EP reduces token costs by 68.64\% on average, offering a scalable, cost-effective solution for budget-constrained reasoning tasks.
\revise{
For resource-rich scenarios, we evaluate TALE-EP under a larger token budget as detailed in \autoref{subsec:larger_budget}.
}

To further evaluate the generalization of \ours{}-EP across different LLMs.
We conduct experiments across Yi-lightning, GPT-4o-mini, GPT-4o and o3-mini on MathBench-College.
\autoref{tab:generalization} illustrates the results, showing \ours{}-EP's ability to reduce output tokens and expenses while maintaining competitive accuracy significantly.
\ours{}-EP achieves substantial token savings, reducing output tokens by 64.63\% on average, compared to Vanilla CoT.
Expense reductions are equally notable, with costs decreasing by 45.30\% on average. 
Despite these cost savings, \ours{}-EP maintains strong accuracy, achieving 76.67\% on Yi-lightning, 70.00\% on GPT-4o-mini, and 80.00\% on GPT-4o, comparable to Vanilla CoT.
These results highlight \ours{}-EP’s effectiveness in balancing cost efficiency and reasoning performance across diverse LLM architectures.
The observed accuracy drop is most significant for GPT-4o-mini. This could be attributed to its smaller number of parameters, which makes it more challenging to answer correctly within a limited response reasoning length.
\revise{We also evaluate the applicability of TALE on more tasks, the results are illustrated in \autoref{subsec:generalization_task}.}
\revise{The efficiency analysis of TALE-EP is in \autoref{subsec:efficiency_tale_ep}
}

\subsection{Effectiveness of \ours{}-PT}
\label{subsec:effective_tale_pt}
\autoref{tab:internalization} compares \ours{}-PT methods with Vanilla CoT and Direct Answering on GSM8K and GSM8K-Zero using Llama-3.1-8B-Instruct. For GSM8K, Direct Answering demonstrates the lowest token usage (38.54) but at the cost of significantly reduced accuracy (21.00\%). In contrast, Vanilla CoT achieves much higher accuracy (77.56\%) but incurs a significant increase in token cost (241.51). 
Note that on GSM8K-Zero, the accuracy of Vanilla CoT drops below Direct Answering.
This drop can be attributed to overthinking, as GSM8K-Zero is simpler, with answers often implied directly within the question. In such cases, a long reasoning process can introduce unnecessary complexity, leading to reduced accuracy.
Among the \ours{}-PT methods, \ours{}-PT-SFT achieves the best accuracy (78.57\%, 78.43\%) with reduced tokens, while \ours{}-PT-DPO balances accuracy (74.11\%, 78.41\%) and token efficiency, cutting token consumption by over 50\% on GSM8K-Zero compared to Vanilla CoT.

%% file: sections/conclusion.tex
\vspace{-0.1cm}
\section{Conclusion}
\vspace{-0.1cm}
\label{sec:conclusion}
In this paper, we introduce \ours{}, a framework that reduces token redundancy in Chain-of-Thought (CoT) reasoning by incorporating token budget awareness.
\ours{} dynamically adjusts the number of reasoning tokens based on the reasoning complexity of each problem, balancing token efficiency and answer correctness.
Experiments show that \ours{} reduces output token usage and expense significantly with acceptable accuracy loss, outperforming Vanilla CoT in cost-effectiveness while generalizing well across various LLMs.

%% file: sections/limitation.tex
\vspace{-0.1cm}
\section{Limitations}
\label{sec:limitations}
\vspace{-0.1cm}
The experiments of our proposed token-budget-aware reasoning framework currently focus on LLMs that process only text as input and output.
While the results demonstrate significant improvements in efficiency and cost reduction, it does not account for models that have multimodal output content. Such as the models generate interleaved images and text as output.
In future work, we will extend token-budget awareness to such LLMs with multimodal output by introducing modality-specific budget constraints and designing adaptive strategies to optimize token efficiency for different modality types, such as images and videos.

%% file: sections/appendix.tex
\appendix

\section{Appendix}
\label{sec:appendix}

\begin{figure}[!ht]
\centering
     \subfloat[Direct answering (10 output tokens).]
    {
       \includegraphics[width=0.83\columnwidth]{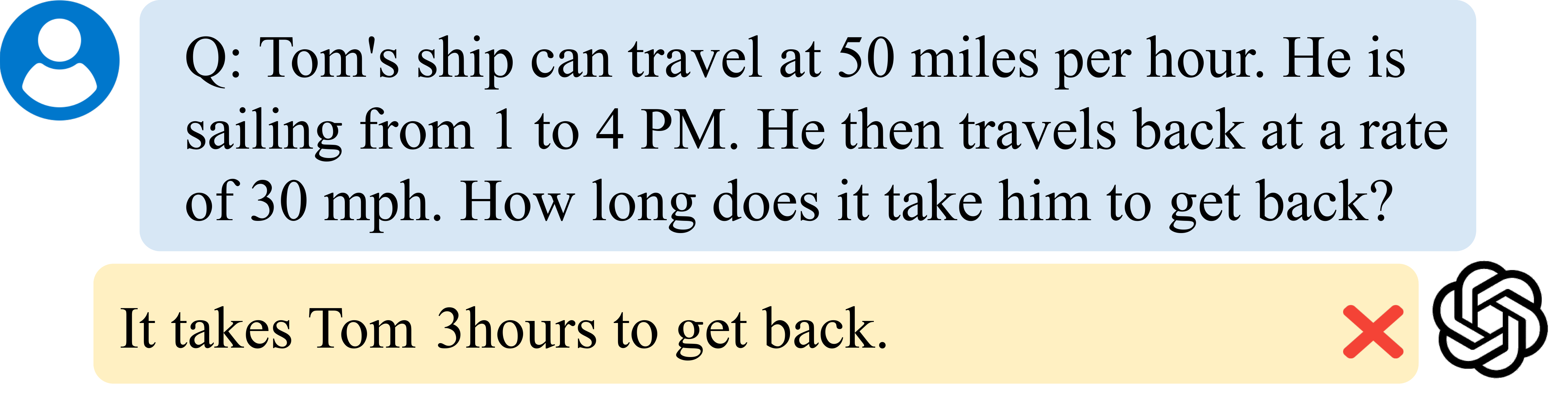}
    }
    \vfill
    \vspace{-8pt}
    \subfloat[Vanilla CoT (271 output tokens).]
    {
        \includegraphics[width=0.83\columnwidth]{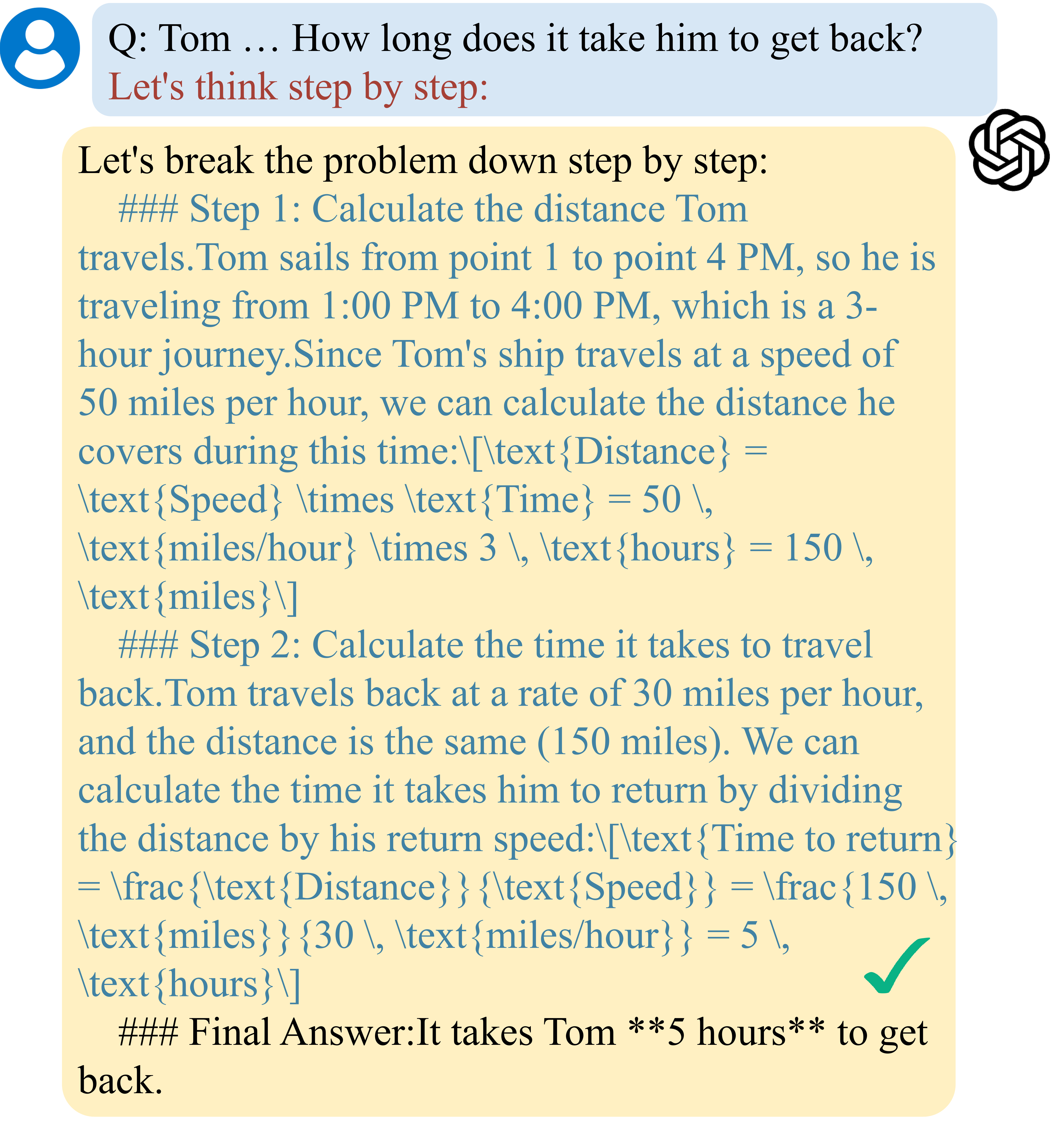}
    }
    \vfill
    \vspace{-8pt}
    \subfloat[\ours{} (68 output tokens).]
    {
        \includegraphics[width=0.83\columnwidth]{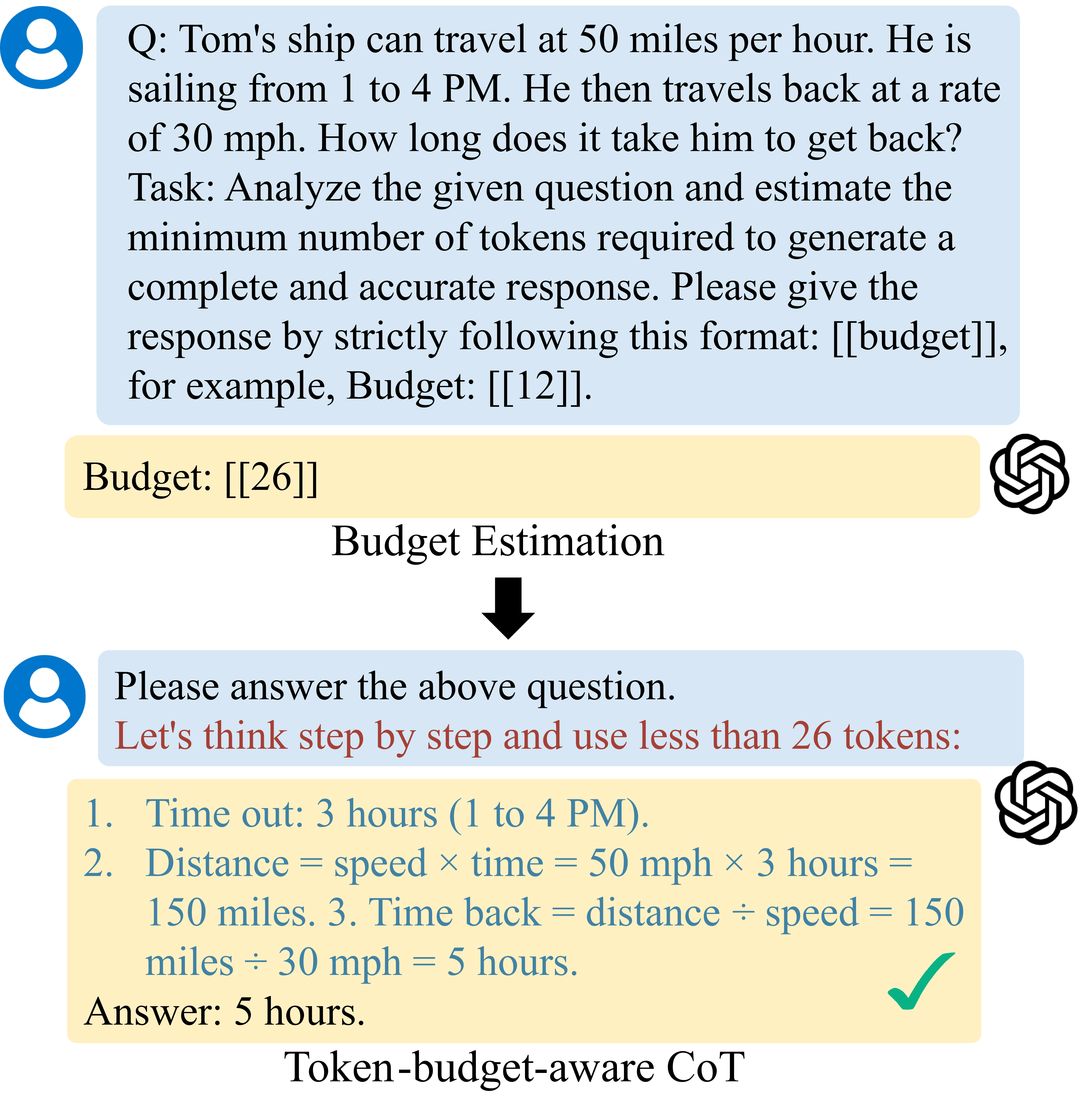}
        \label{fig:intuitive_example_workflow_ep}
    }
    \vspace{-6pt}
    \caption{An intuitive example to illustrate the workflow of \ours{}-EP on GPT-4o-mini~\cite{gpt4o-mini2024}.}
    \label{fig:intuitive_example_workflow}
    \vspace{-16pt}
\end{figure}

\begin{figure}[]
    \centering
    \includegraphics[width=0.83\linewidth]{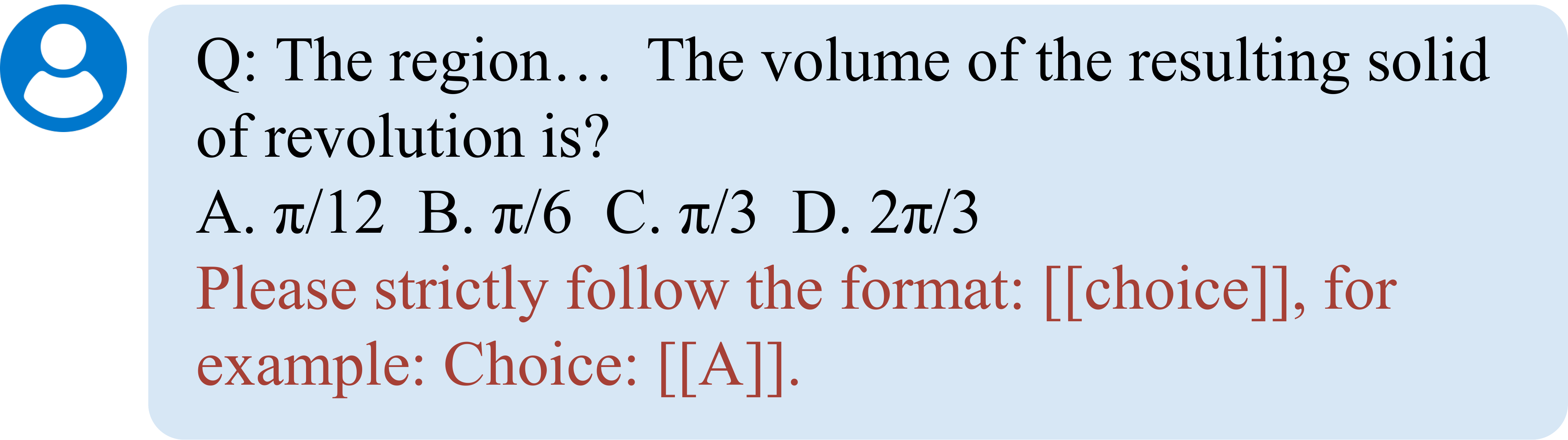}
    \vspace{-0.1cm}
    \caption{The instruction prompt used to format the LLM output on multiple-choice questions.}
    \label{fig:format_prompt}
    \vspace{-0.4cm}
\end{figure}

\subsection{Definition of Ideal Budget Range}
\label{subsec:problem_formulation}
\textbf{Ideal Budget Range.}
Based on the observation of token elasticity, a token cost bottom range exists during searching for the optimal budget.
In this range, the token costs approach the token cost lowest bound.
Before or after the range, the token cost will increase.
We define such a bottom range as ``ideal budget range''.
It's worth noting that the budget continuously degrades during the search. Only the token cost rebounds. That's why we refer to this observation as token elasticity.
To summarize, ideal budget range is an range that minimizes actual token consumption.
Let $\boldsymbol{\beta} = \{\beta_1, \beta_2, ..., \beta_N\}$ denote all possible budgets that can maintain answer correctness.
A rolling window $W \in \boldsymbol{\beta}$ is applied iteratively over $\boldsymbol{\beta}$.
Let $k$ represent the range size, which is adaptively determined during our evaluation as $\frac{N}{3}$, where $N$ is the total number of possible budgets.
A budget range is defined as:
\begin{equation*}
\begin{aligned}
    W_k(i) = \{\boldsymbol{\beta}_j \mid i \leq j \leq i + k - 1\}, 
    \\
    1 \leq i \leq |\boldsymbol{\beta}| - k + 1
\end{aligned}
\end{equation*}
The ideal budget range $W^*$ is defined as:
\begin{equation}
   W_k^* = \arg \min_i \left( \sum_{\beta_j \in W_k(i)} \mathbb{T}(\beta_j) \right),
    \label{eq:ideal_budget_interval}
\end{equation}
where $\mathbb{T}$ denote the actual token consumption for a given budget $\beta \in \boldsymbol{\beta}$.
We aim to estimate a budget located in the ideal budget ranges without any search process.
In that case, \ours{} obtains the ideal budget within acceptable sacrifice.

\subsection{Effectiveness of Budget Estimation.}
\label{subsec:rq2_budget_estimation}
\begin{figure}[!t]
\centering
     \subfloat[Successful case.]
    {
       \includegraphics[width=0.83\columnwidth]{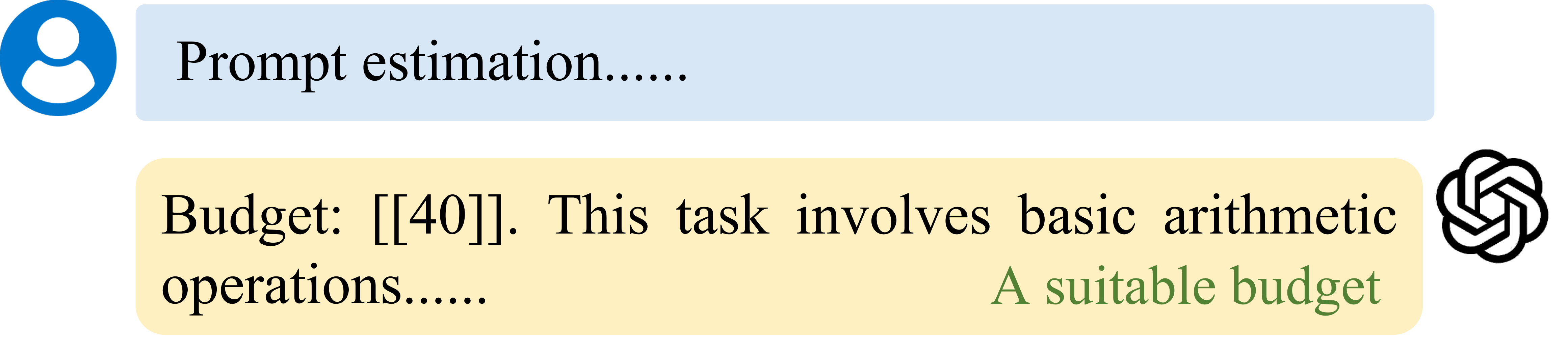}
    }
    \vfill
    \vspace{-8pt}
    \subfloat[Failed case.]
    {
        \includegraphics[width=0.83\columnwidth]{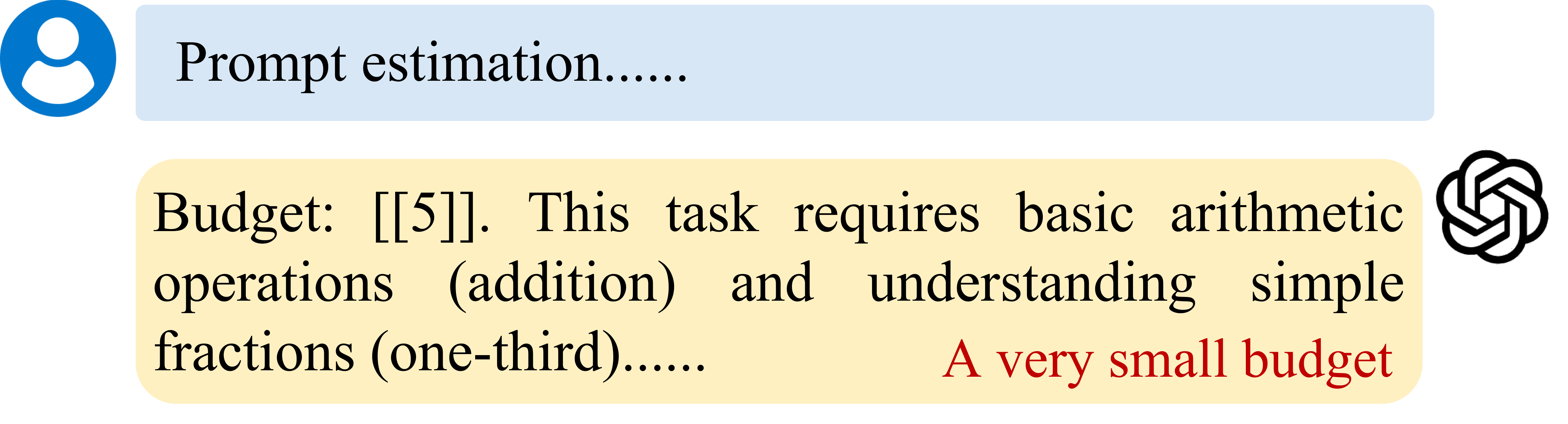}
    }
    \vspace{-6pt}
    \caption{\revise{An intuitive example for successful and failed cases of prompt budget estimation in TALE-EP.}}
    \label{fig:successful_fail_case_EP}
    \vspace{-8pt}
\end{figure}

In this RQ, we evaluate the effectiveness of the budget estimation performance.
An ideal estimated budget should be located around the optimal searched budget and in the bottom area of \autoref{fig:motivation_elastic_observation}.
We further define such an area as the ideal budget range and give the formalized definition in \autoref{subsec:problem_formulation}.
A good budget should be located in the ideal budget range.
Two metrics are taken into consideration: \textit{in-range accuracy} and \textit{out-of-range distance}.
In-range accuracy determines whether the predicted budget $\hat{\beta}$ falls within the ideal budget range $W^{*}_{k}$. Mathematically, it can be expressed as:
\begin{equation*}
    \mathbb{I}\{\hat{\beta} \in W_k^*\} =
\begin{cases}
1, & \text{if } \hat{\beta} \in W_k^*, \\
0, & \text{otherwise}.
\end{cases}
\end{equation*}
Out-of-range distance quantifies the distance between $\hat{\beta}$ and $W_k^*$ if the predicted budget $\beta^*$ falls outside the ideal budget range $W_k^*$.
Let $dist(\hat{\beta}, W_k^*)$ represent the distance, defined as:
\begin{equation*}
    \text{dist}(\hat{\beta}, W_k^*) =
\begin{cases}
0, & \text{if } \hat{\beta} \in W_k^*, \\
\min\limits_{\substack{\hat{\beta} \in W_k^*}} |\hat{\beta} - \beta|, & \text{if } \hat{\beta} \notin W_k^*.
\end{cases}
\end{equation*}
Intuitively, a higher in-range accuracy and a lower out-range distance indicate a better estimated budget. 
During our evaluation, the in-range accuracy is 60.61\%, and the out-of-range distance is 109.64.
It indicates that more than two-thirds of estimated budgets are located in the ideal range.
For those out-of-range samples, they have an offset of 109.64 tokens on average.
\autoref{fig:successful_fail_case_EP} illustrates the successful and failed estimated cases intuitively. 
\revise{
The prompt we use for budget estimation is as follows:

\texttt{``Task: Analyze the given question and estimate the minimum number of tokens required for reasoning.''}

This prompt encourages the model to evaluate the question as a whole, including reasoning depth, structure, completeness, and surface-level difficulty. 
}

\subsection{Details of \ours{}'s Implementation}
\label{subsec:details_implementation}
In this section, we introduce the hyper-parameters used for \ours{}-EP and \ours{}-PT.

\noindent\textbf{\ours{}-EP.}
\ours{}-EP uses a zero-shot mechanism to estimate the token budget and then prompts the LLM. The instruction prompts used during this process are shown in \autoref{fig:intuitive_example_workflow}. To ensure output consistency, we set the temperature to 0.1 and limit the model to a single reasoning path. Additionally, the random seed is fixed at 1024.

\noindent\textbf{\ours{}-PT.}
\ours{}-PT includes two implementations: SFT and DPO. For parameter efficiency, both implementations adopt LoRA~\cite{hu2021lora} for post-training, with rank set to 8 and lora alpha set to 32. 
For \ours{}-PT-SFT, we train for 3 epochs with a batch size of 16, a learning rate of 1e-4, and a weight decay of 0.01. For \ours{}-PT-DPO, we train for 2 epochs with a batch size of 16, a learning rate of 3e-5, and a weight decay of 0.001.

\subsection{Comparison of TALE-EP and TALE-PT.}
\label{subsec:TALE-EP-TALE-PT}
\revise{
In this section, we compare the performance of TALE-EP and TALE-PT (including SFT and DPO).
Specifically, we utilize Llama-3.1-8B-Instruct as both the budget estimator for TALE-EP and the base model for TALE-PT, evaluated on GSM8K.
\autoref{tab:comparison_tale-ep_tale-pt} illustrates the evidence.
As the zero-shot-based estimator tends to predict relatively low budgets, TALE-EP achieves lower token usage but at the cost of slightly reduced accuracy.
In contrast, both variants of TALE-PT achieve higher accuracy with more tokens, as their training data is constructed using optimal budget search, which enforces answer correctness as a strict constraint even with a higher token costs.
This highlights a trade-off between strict correctness preservation (in TALE-PT) and token efficiency (in TALE-EP). 
}
\begin{table}[!t]
    \centering
    \scriptsize
    \tabcolsep=5pt
    \renewcommand{\arraystretch}{1.2}
    \caption{Comparison of TALE-EP and TALE-PT.}
\vspace{-0.2cm}
\begin{tabular}{cccc}
\toprule
\multirow{2}{*}{Metrics} & \multirow{2}{*}{TALE-EP} & \multicolumn{2}{c}{TALE-PT} \\
\cmidrule(lr){3-4}
 &  & SFT & DPO \\
 \midrule
ACC & 71.82 & 78.57 & 74.11 \\
Output Tokens & 112.21 & 139.63 & 149.93 \\
\bottomrule
\end{tabular}
\label{tab:comparison_tale-ep_tale-pt}
 \vspace{-0.3cm}
\end{table}

\subsection{Applicability of \ours{} on More Tasks.}
\label{subsec:generalization_task}
\revise{
To further evaluate the applicability, we deploy TALE-EP on three additional open-ended generative tasks. As these tasks involve open-ended text generation, we adopt the BLEU metric~\cite{papineni2002bleu} to quantify the similarity between generated outputs and reference texts. The results in \autoref{tab:generalization} demonstrate that TALE-EP achieves comparable or even better BLEU scores than Vanilla CoT while using only around 40\% of the output tokens, validating its effectiveness and applicability in broader generative tasks.
}

\begin{table}[!t]
    \centering
    \scriptsize
    \tabcolsep=5pt
    \renewcommand{\arraystretch}{1.2}
    \caption{Generalization of \ours{} on more tasks. Three popular LLM generative tasks, Code Summarization~\cite{husain2019codesearchnet}(CS), Empathetic Response Generation~\cite{rashkin2019towards}(ERG), Code Generation~\cite{austin2021program}(CG), are taken into consideration. BLEU is taken as the metric to evaluate the performance. BLEU$\uparrow$. Output Tokens$\uparrow$.}
\vspace{-0.1cm}
\begin{tabular}{lcccc}
\toprule
\multirow{2}{*}{Tasks} & \multicolumn{2}{c}{TALE-EP} & \multicolumn{2}{c}{Vanilla CoT} \\
\cmidrule(lr){2-3} \cmidrule(lr){4-5} 
 & BLEU & Output Tokens & BLEU & Output Tokens \\
 \midrule
CS & 0.07 & 44.39 & 0.2 & 134.05 \\
ERG & 0.005 & 60.34 & 0.006 & 175.37 \\
CG & 0.24 & 171.08 & 0.267 & 461.77 \\
\bottomrule
\end{tabular}
\label{tab:more_tasks}
 \vspace{-0.3cm}
\end{table}

\subsection{Formalizing the Budget Search.}
\label{subsec:formalizing_budget_search}
\revise{
For a given input $\vx$, we define the search space as:
\begin{equation*}
    \mathcal{B} = \{ \beta \in \mathbb{Z}^+ \mid 0 < \beta \leq T_{\text{vanilla}}(x) \}
\end{equation*}
where $T_{\text{vanilla}}(x)$ is the number of tokens generated by Vanilla CoT.
A candidate budget $\beta \in \mathcal{B}$ is considered feasible if:
\begin{equation*}
    LLM(x,\beta)=y \text{ and } T(x,\beta)<T(x,\beta_0)
\end{equation*}
where $y$ is the ground-truth answer, $T(x, \beta)$ is the actual number of output tokens when answering $x$ under budget $\beta$, $\beta_0$ is the previously searched (larger) feasible budget.
Our goal is to find:
\begin{equation*}
    \beta^{*}=\arg \min _{\beta \in \mathcal{B}} T(x, \beta) \text{, subject to  } LLM(x, \beta)=y
\end{equation*}
To efficiently find $\beta^*$, we employ a binary search procedure guided by the feasibility function above, as detailed in \autoref{alg:binary_search} and \autoref{alg:greedy_feasibility}.
}

\subsection{Efficiency of TALE-EP.}
\label{subsec:efficiency_tale_ep}
\revise{
Since TALE-EP requires one additional query, we further evaluate its end-to-end latency in this section.
Specifically, we query the Llama-3.1-8B-Instruct model over the GSM8K-Zero dataset and measure both accuracy and average time cost. T
he budget estimation query of TALE-EP is also considered.
Although TALE-EP requires one additional query compared to Vanilla CoT, it is significantly more efficient, taking only 2.3 seconds per sample, while Vanilla CoT takes 10.2 seconds. This is because the primary factor influencing inference time is the number of output tokens, which TALE-EP effectively reduces.
}

\subsection{Effectiveness of Larger Token Budget.}
\label{subsec:larger_budget}
\revise{
In scenarios with ample computational resources, the token budget could be larger for better performance.
We simulate such a scenario by scaling the estimated budget by a factor $\alpha$ ($\alpha * \text{budget}$). As shown in the table below, increasing $\alpha$ from 1 to 2 leads to higher accuracy (from 67.33\% to 72.66\%) at the cost of more tokens (from 210.97 to 279.78), demonstrating that TALE-EP can flexibly adapt to different resource scenarios.
}

\begin{table}[!t]
    \centering
    \scriptsize
    \tabcolsep=5pt
    \renewcommand{\arraystretch}{1.2}
    \caption{The empirical evidence for ``implicit monotonicity assumption''. $\bar{\beta}$ is the budget upper bound, which is the token cost of vanilla CoT. The budget row displays scaled budgets ranging from $2^{-2}$ to $2^{2} \cdot \beta^*$.}
\vspace{-0.2cm}
\begin{tabular}{lcccc}
\toprule
\textbf{Budget($*\beta^*$)} & $2^{-5}$ & $2^{-4}$ & $2^{-2}$ & $2^{0}$  \\ 
\midrule
\textbf{ACC} & 69.23 & 75.82 & 75.82 & 76.92  \\ 
\midrule
\textbf{Output Tokens} & 222.69 & 222.42 & 244.61 & 653.53  \\ 
\bottomrule
\end{tabular}
\label{tab:assumption_evidence}
 \vspace{-0.5cm}
\end{table}

\subsection{Empirical Evidence for the ``Implicit Monotonicity Assumption''.}
\label{subsec:assumption_evidence}
\revise{
In this section, we give empirical evidence to support the ``implicit monotonicity assumption'' for our search algorithm.
Specifically, for a set of questions, we vary the budget across a range of values (e.g., $2^{-5}, 2^{-4}, 2^{-2}, 2^0$ times budget upper bound, which is the token cost of vanilla CoT), and record the corresponding accuracy and average output tokens at each point. 
\autoref{tab:assumption_evidence} illustrates the empirical evidence.
Observe that it roughly follows from the monotonicity property. 
The results demonstrate a consistent trend: accuracy generally increases or plateaus with increasing budgets, confirming a soft monotonicity in most cases.
}